\documentclass[journal]{IEEEtran}
\usepackage{cite}
\usepackage{amsmath,amssymb,amsfonts}
\usepackage{algorithmic}
\usepackage{graphicx}
\usepackage{textcomp}
\usepackage{graphics} 
\usepackage{epsfig} 
\usepackage{listings} 
\usepackage{multirow}
\usepackage{longtable}
\usepackage{graphicx}
\usepackage{textcomp}
\usepackage{rotating}
\usepackage{pdflscape}
\usepackage{afterpage}     
\usepackage{verbatim}
\usepackage{dcolumn}
\usepackage{adjustbox}
\usepackage[table,xcdraw]{xcolor}
\usepackage{url}
\usepackage[flushleft]{threeparttable}
\usepackage[caption=false]{subfig}
\usepackage{graphicx}
\usepackage{caption}
\usepackage{lipsum} 
\usepackage{array,multirow}
\def\BibTeX{{\rm B\kern-.05em{\sc i\kern-.025em b}\kern-.08em
    T\kern-.1667em\lower.7ex\hbox{E}\kern-.125emX}}
\begin{document}

\title{Qualitative and Quantitative Risk Analysis and Safety Assessment of Unmanned Aerial Vehicles Missions over the Internet\thanks{The paper is accepted for publication in IEEE Access, April 2019.}}

\author{Azza~Allouch,
        Anis~Koub\^aa,
				Mohamed~Khalgui, 
        and~Tarek~Abbes
\thanks{A. Allouch is with the School of Intelligent Systems Science and Engineering, Jinan University (Zhuhai Campus), Zhuhai 519070, China, also with the Faculty of Mathematical, Physical and Natural Sciences of Tunis , University of Tunis El Manar, Tunis 1068, Tunisia, and also
with the National Institute of Applied Sciences and Technology, University
of Carthage, Tunis 1080, Tunisia (email: azza.allouch@coins-lab.org).}
\thanks{A. Koub\^aa is  is with the Department of Computer Science, Prince Sultan
University, Riyadh PSU 12435, Saudi Arabia, also with CISTER/INESC TEC,
4200-135 Porto, Portugal, and also with ISEP-IPP, Porto, Portugal, (email: akoubaa@coins-lab.org).}
\thanks{M. Khalgui is with the School of Intelligent Systems Science
and Engineering, Jinan University (Zhuhai Campus), Zhuhai 519070,
China, and also with the National Institute of Applied Sciences
and Technology, University of Carthage, Tunis 1080, Tunisia (e-mail:
khalgui.mohamed@gmail.com).}
\thanks{T. Abbes is with the National School of Electronics and Telecommunication, Sfax 3018, Tunisia, (email: tarek.abbes@enetcom.usf.tn).}}

\markboth{Accepted in IEEE Access, April~2019}%
{Shell \MakeLowercase{\textit{et al.}}: Bare Demo of IEEEtran.cls for IEEE Journals}

\maketitle

\begin{abstract}
In the last few years, Unmanned Aerial Vehicles (UAVs) are making a revolution as an emerging technology with many different applications in the military, civilian, and commercial fields. The advent of autonomous drones has initiated serious challenges, including how to maintain their safe operation during their missions. The safe operation of UAVs remains an open and sensitive issue, since any unexpected behavior of the drone or any hazard would lead to potential risks that might be very severe. The motivation behind this work is to propose a methodology for the safety assurance of drones over the Internet {(Internet of drones (IoD))}. Two
approaches will be used in performing the safety analysis: (1) a qualitative safety analysis approach, and (2) a quantitative safety analysis approach. The first approach uses the international safety standards, namely ISO 12100 and ISO 13849 to assess the safety of drone's missions by focusing on qualitative assessment techniques.
The methodology starts from hazard identification, risk assessment, risk mitigation, and finally draws the safety recommendations associated with a drone delivery use case. The second approach presents a method for the quantitative safety assessment using Bayesian Networks (BN) for probabilistic modeling. BN utilizes the information provided by the first approach to model the safety risks related to UAVs' flights.
An illustrative UAV crash scenario is presented as a case study, followed by a scenario analysis, to demonstrate the applicability of the proposed approach. These two analyses, qualitative and quantitative, enable { all involved stakeholders} to detect, explore and address the risks of UAV flights, which will help the industry to better manage the safety concerns of UAVs.\end{abstract}


%
\IEEEpeerreviewmaketitle

\section{Introduction}
\label{sec:introduction}
Unmanned {Aerial Vehicles (UAVs) have become an extremely popular technology \footnote{(2018) TopDr.One, ``DRONE SALES STATISTICS
'' [Online]. Available: https://topdr.one/drone-sales-statistics/
.}. These flying robots have promoted the development of several applications such as sensing, smart cities, surveillance \cite{benjdira2019car}, disaster management and recovery, patrolling, aerial survey, and border security \cite{pajares2015overview}. The drone technology is becoming increasingly popular, and UAVs are anticipated to be even more widely adopted in the future. This is confirmed by the Federal Aviation Administration (FAA) in the US, which expects that the number of UAVs consumers will increase from 1.1 million to 3.55 million between 2016 and 2021 \footnote{(2018) William Atkinson, ``Drones Are Gaining Popularity'' [Online]. Available: https://www.ecmag.com/section/your-business/drones-are-gaining-popularity.}.}

{The advent of autonomous UAVs offers significant benefits; however, there are still open challenges that restrain their real-world deployment. Because of the limited computational resources of these low cost drones, their integration with the Internet-of-Things and the cloud is an emerging trend. There have been a very few attempts to integrate drones with the Internet
and IoT \cite{gharibi2016internet}. In \cite{koubaa2017service,KOUBAA201946,qureshi2016poster}, the authors developed a cloud robotics platform, Dronemap Planner (DP), that allows the monitoring, communication and real-time control of robots and drones over the Internet. Dronemap integrates UAVs with the cloud and aims to virtualize the access to UAVs, and offload heavy computations from the UAVs to the cloud. In \cite{koubaa2018dronetrack}, the authors proposed a cloud system for real-time monitoring of multi-drone systems used for the tracking of moving objects. The advent of autonomous drones has initiated serious challenges like safety and security \cite{schenkelberg2016reliable}. 

For this purpose, the safety of drones operations must be ensured, as part of non-functional properties of the system \cite{erceg2017unmanned}. Safety can be defined as a ``state in which the system is not in danger or at risk, free of injuries or losses''\cite{sanz2015safe}. Because nothing is totally safe and there is no situation where no risk can occur, safety is also defined as the absence of unacceptable risks\cite{ISO50,jespen2016risk}. }

{In fact, the use of civilian drones is still in its infancy \cite{custers2016future}\cite{aerosystems1999civilian}. In addition, UAVs are special categories of cyber-physical systems that communicate using wireless, which makes them more prone to safety risks and security threats \cite{Safe2018}. Since this is a relatively new emerging field, hazards and risks of UAV flights are still not completely known nor understood, which {might jeopardize the safety of UAV missions, especially} due to the absence of standards and regulations that govern the safe use and operation of UAVs\cite{waibel2017drone}. This is the main reason behind the limited utilization of drones {for} civilian purposes {\footnote{(2011) Unmanned Aircraft Systems (UAS). [Online]. Available: https://www.trade.gov/td/otm/assets/aero/UAS2011.pdf}}
 \cite{rajagopalan2018drones}.}

{Moreover}, the lack of policies, standards and guidelines that govern the safe use, operation and emerging safety problems of civilian drones creates a significant barrier to research and development\cite{watts2012unmanned}. The limits of today\textquotesingle s approaches to safety have also been clearly demonstrated by the {growing} list of safety accidents and incidents\cite {waibel2017drone}. According to FAA {\footnote{(2017) Civilian Drone Safety Incidents Keep Rising. [Online]. Available: https://www.insurancejournal.com/news/national/2017/12/08/473529.htm}} , more than 4,889 incidents have been reported between 2014 and 2017, which can easily inflict serious {harm} to people or properties.

{For this reason, safety should be considered as {a core} requirement in every system or application design, and even more for systems that can provoke great damages \cite{sanz2015safe}. Thus, without a clear understanding of potential risks of usage of these drones, the public use of civilian drones will not be possible at an acceptable safety level.}

\subsection{Problem Statement and Motivation}
The safety assessment is an initial step to regulate the safe use of UAVs. {Different strategies have been proposed with respect to the safety assessment of UAVs, i.e. assessing the risk of an undesired event in a system, using both quantitative \cite{GONCALVES2017383,barr2017preliminary} and qualitative \cite{mhenni2016safesyse,sanz2015safe,neff2016identifying,sankararaman2017towards,NAP25143} approaches}. For instance, the FAA developed a pre-flight assessment process in which the Safety Management System (SMS) \footnote{(2014) FAA Administration, ``Sms safety management system
manual version 4.0,'' [Online]. Available: https:
//www.faa.gov/airports/airport safety/safety management systems/
external/pilot studies/documentation/jqf/media/jqfSMSdraftmanual.pdf.} was adopted to identify the risks and the risk mitigation strategies to ensure that no safety hazards will occur during the UAV mission. In \cite{mhenni2016safesyse}, Mhenni et al. recommended a framework for regulating the safety of UAV operations, including a Model-Based Systems Engineering (MBSE) and a Model-Based Safety Analysis (MBSA), integrating quantitative techniques such as Fault Tree Analysis (FTA) and qualitative techniques such as Failure Mode and Effects Analysis (FMEA).

Gon\c{c}alves et al. \cite{ GONCALVES2017383} {presented a safety assessment process model for a UAV using Petri Nets, while \cite{barr2017preliminary} used Bayesian Belief Networks for performing risk analysis of small unmanned aircraft systems}.

{Some works such as \cite{ neff2016identifying} \cite{sankararaman2017towards} proposed the identification and assessment of UAV risk-factors  (as obstacle collision, untimely battery drain, human factor) based on qualitative analysis}. The current FAA Order 8040 approach to risk management is also based on fundamentally qualitative and subjective risk analysis \cite{NAP25143}. The qualitative nature of the current approach might lead to results that fail to be repeatable, predictable, and transparent.

{On the other hand, a rigorous safety assessment also requires a quantitative analysis \cite{ bjorkman2011probabilistic}. This is even confirmed by \cite{NAP25143}, which recommends establishing quantitative probabilistic risk analyses. Therefore, for completeness purposes, we concluded that it is necessary to conduct detailed studies on the UAV safety assessment at both qualitative and quantitative levels. In this paper, we follow a combination of qualitative and quantitative analyses to identify hazards and risks that can occur when drones are tele-operated over the Internet (IoD). We propose safety procedures, safeguards and protective measures to reduce the risk to an acceptable level. We also investigate the main reasons that might lead to drone crashes, and analyze them in order to identify mitigation measures that must be taken into account to avoid crashes and accidents. Moreover, we identify probabilistic metrics of drone crash events given some states of the system.}

\subsection{Approach and Contributions}

To address this problem, we propose two approaches: The first approach is qualitative and is based on the functional safety standards ISO 12100 and ISO 13849. The second approach is quantitative and is based on Bayesian Networks. 
\subsubsection{Baseline Standards and Approaches}

The new Machinery Directive 2006/42/EC requires a risk assessment for a machine. ISO 12100 \footnote{(2010) Safety and functional safety a general guide. [Online]. Available:
https://library.e.abb.com/public/.../1SFC001008B0201.pdf.} is a type A standard that applies to everything that is defined as a machine under the European Machinery Directive. It is used for machines for which there is no type C standard, i.e. no standard dedicated to the specific product or machine under consideration \cite{iso201012100}. {In our case, there is no type C standard dedicated to the drones, thus we propose to adapt ISO 12100 for the functional safety assessment of drones}. This standard specifies the basic terminologies and principles of the risk assessment and the risk reduction for ensuring safety in machinery design. These principles are based on the knowledge and experience, past accidents and incidents, and hazards associated with the machinery.

To comply with the machinery directives requirements, { the harmonized standard ISO 13849 \footnote{(2018) Machinery Directive \& Harmonised Standards. [Online]. Available:
https://www.cem4.eu/component/attachments/download/501.}} is the most relevant from the functional safety point of view as compared to {IEC 61508 \footnote{(2017) Functional safety, Technical guide No. 10. [Online]. Available:
https://library.e.abb.com/public/acd23f92341a4d50bf3500a245494af8/
EN\_TechnicalguideNo10\_REVF.pdf.}}, which is not a harmonized European standard. It cannot be used as a proof of the CE conformity {(European Conformity). It is to be noted that a harmonised standard is an European standard that demonstrates how a product or a machine complies with the European Conformity. }
The ISO 13849 standard is internationally recognized and is a Performance Level (PL)-focused standard whose outcomes can be equated to IEC Safety Integrity Level (SIL) standards, which makes it even more useful. ISO 13849 is a standard that can cover most, if not all, concerns of the machine manufacturer {(OEM: Original Equipment Manufacturer) \footnote{Understanding Machine Safety Guidelines. [Online]. https://www.festo.com/rep/en\-us\_us/assets/pdf/FESTO\_eGuide\_final2.pdf.}} in factory automation safety controls . It combines the complex probability method from IEC 61508 and the deterministic category approach from EN 954-1 based on the risk assessment.

Unlike many of the other international standards, ISO 13849 is used for the safety-related parts of control systems and diverse types of machinery; it applies to all technologies, while {IEC 61511\cite{bond2002iec}} is specific for the process industries and {IEC 62061\cite{iec2005safety}} can only be applied to electronic components.

{Most functional safety assessments are performed in a qualitative manner, as mentioned in the aforementioned standards. However, for the sake of comprehensiveness of the safety analysis, it is also important to complement the qualitative analysis by a quantitative analysis.} 

Quantitative analysis uses model-based techniques for probability estimation, which allows systems to be analysed in a more formal way. {The Bayesian Networks (BN) formalism is a commonly used approach for quantitative risk assessment \cite{jun2017bayesian,kwag2018probabilistic,brito2016bayesian,zhang2014bayesian}. It captures the relations between faults and symptoms, identifies and estimates the probabilities of risks in various scenarios. For this reason, we will use Bayesian Networks as our quantitative analysis approach for drone's safety analysis. }

\subsubsection{Our Methodology}

Our approach consists {in conducting a} comprehensive study of safety aspects for the usage of civilian drones at a public scale through the Internet. We combine the ISO 13849 and ISO 12100 standards to derive a unified functional safety methodology. In addition, a process is proposed for identifying and classifying hazardous conditions, along with their possible causes and their consequences that affect the safe operation of UAVs. Furthermore, the risk mitigation strategies required to reduce the associated risk to an acceptable level are outlined to ensure the safety of drone's missions when tele-operated and monitored through the Internet. Based on the output of the first approach, a quantitative evaluation method based on Bayesian Networks is then presented in this paper. The Bayesian model of the UAV crash risk is developed by causality, and the expected probabilities of crashes associated with an UAV system is analysed based on real flight data derived from literature {\cite{fernando2017survey,belcastro2016aircraft,clothier2015safety,boyd2015causes,enomoto2013preliminary}}. Finally, an illustrative example is demonstrated, and the simulation results are discussed.

In summary, the main contributions of this paper are as follows:
\begin{itemize}
	\item First, we propose a qualitative safety analysis for functional safety for drone crashes based on two safety standards, ISO 12100 and ISO 13849.
\end{itemize}
\begin{itemize}
	\item {Second, a drone delivery use case is presented as a case study in order to verify the applicability and the feasibility of the proposed {functional safety methodology.}}
\end{itemize}
 \begin{itemize}
	 \item Third, we perform a quantitative safety analysis using Bayesian Networks based on information given by ISO 12100 and ISO 13849 followed by an illustrative example for demonstration; finally, simulation results are discussed.

 \end{itemize}
{The paper is organized as follows. Section II reviews the recent regulations and research papers related to drones' safety. The qualitative safety approach is proposed in Section III and demonstrated by a drone delivery use case. The Bayesian Networks model is proposed as a quantitative safety analysis method to validate the proposed functional safety methodology followed by an UAV scenario analysis to demonstrate its feasibility. Finally, Section IV provides some concluding remarks and suggestions for a future work.}

\section{Related Works}
The emerging drone industry is greatly evolving, yet still in the process of establishing safety standards. To date, no safety standard exists for autonomous drone systems. European regulations are still being drafted \cite{waibel2017drone} and will be centered on the safety risks posed by drones. Due to the current absence of international standards, laws and guidelines that govern the safe use and operation of civilian drones, many groups, including companies and researchers, have issued their own framework for regulating the safety of UAV operations.

\subsection{UAV regulatory organizations and authorities }

For maintaining the safety of UAVs and the public, the FAA in the United States has developed rules to regulate the use of small UAVs. The FAA has put out a ``Know Before You Fly program'' \footnote{Know before you fly. Accessed: 19 January 2017. [Online]. Available: www.knowbeforeyoufly.com.} to provide guidance on the responsible use of UAV and educate the public about UAV safety. The FAA requests a preflight assessment with risk mitigation strategies to ensure that the UAV will pose no risk to aircraft, people, or property when the UAV loses control or other safety hazards.

The International Civil Aviation Organization (ICAO) is an international organization that collaborates with national civil aviation authorities. It is concerned with basic regulatory frameworks and provides information and guidelines for Air Navigation Services. In 2016, the ICAO published an online toolkit \footnote{www.icao.int/rpas.} that provides general guidelines for regulators and operators \cite{ICAO} . The same organization further issued recommendations for the safe integration of UAVs into controlled airspace.

The Joint Authorities for Rule-making on Unmanned Systems (JARUS) is a group of national authorities that aims to provide guidance to support and facilitate the creation of UAV regulations. In particular, they recommend regulations that focus on UAV safety. Further, Euro-control and the organization JAA (Joint Aviation Authority) created the UAV Task Force, with the goal of integrating UAVs in European airspace through setting a guiding report on UAV safety requirements.
In March 2015, the European Aviation Safety Agency (EASA) published a regulatory approach for UAVs, called ``the Concept of Operations for Drones: A risk-based approach to regulation of unmanned aircraft'' \cite{EASA}, which focused on the integration of drones into the existing aviation system in a safe manner. This was followed by the publication of the EASA Technical Opinion \cite{EASA2} introducing a Prototype Regulation on Unmanned Aircraft Operations, published in Summer 2016.

{The FAA and EASA together with EUROCAE, USICO, JARUS, ICAO, and UVS International have defined formal policies for UAV certification and a clear regulation for the National Air Space (NAS) management. However, the new regulations proposals will be officially published when EU and US adopts it. These recommended regulations are available and used as common guidelines until the legislation publication. }

\subsection{International safety standards}

 In Europe, to commercialize a machine (robot), it is required to get a CE certification in accordance with the European Directive on machinery 2006/42/EC \cite{directive2006directive}, which states that risk-management techniques should be achieved.

There are two alternative standards bodies that can be followed when implementing functional safety systems in compliance with the Machinery Directive: The International Organization for Standardization (ISO) standard and the International Electrotechnical Commission (IEC) standard \footnote{(2010) ABB brochure, ``Safety and functional safety: A
general guide,'' Tech. Rep., [Online]. Available:
https://library.e.abb.com/public/acd23f92341a4d50bf3500a245494af8/
EN TechnicalguideNo10 REVF.pdf.}.

Considering the nature of UAVs as vehicles or mobile machines, the ISO/IEC standards are highly recommended as they give confidence to the regulatory bodies to deliver certification that ensures compliance with relevant regulations, helps to protect the public, and can be adapted to be applied to UAV outdoor missions. The main functional safety standards in current use are listed below:

\begin{itemize}
	\item IEC 61508: Functional safety of electrical/electronic/programmable electronic safety-related systems \cite{standard201061508}.
\end{itemize}
\begin{itemize}
	\item ISO 61511: Functional safety, safety instrumented systems for the process industry sector \cite{bond2002iec}.
\end{itemize}
\begin{itemize}
	\item EN ISO 13849: Safety of machinery, safety-related parts of control systems, General
principles for design \cite{iso2008safety}.
\end{itemize}
\begin{itemize}
	\item EN 954-1: Safety of machinery, safety-related parts of control systems, General principles for
design \cite{din1996954}.
\end{itemize}
\begin{itemize}
	\item EN 62061: Safety of machinery, functional safety of safety-related electrical, electronic and programmable electronic
control systems \cite{international2005iec}.
\end{itemize}
\begin{itemize}
	\item ISO 26262: Road Vehicles functional safety standard \cite{iso26262}. 
\end{itemize}

Every standard for functional safety-related control systems requires a risk assessment. It is mandatory to perform a risk assessment for a machine according to the new Machinery Directive 2006/42/EC \footnote{(2010) Safety and functional safety a general guide. [Online]. Available:
https://library.e.abb.com/public/.../1SFC001008B0201.pdf.}. Basic safety standards for risk assessment include:
\begin{itemize}
	\item ISO 12100: Safety of machinery-general principles for design-risk
assessment and risk reduction \cite{iso201012100}.
\end{itemize}
\begin{itemize}
	\item ISO 14121: Safety of machinery, risk assessment \cite{iso20}.

\end{itemize}
\begin{itemize}
	\item ISO 31000: Risk management-principles and guidelines \cite{iso200931000}. 
\end{itemize}

\subsection{Risk analysis approach for UAV operations}

In both national and international standards, the risk assessment approach is an initial step to regulate the safe use of UAVs. Several approaches have been proposed in the field of safety assessment of UAVs. In \cite{mhenni2016safesyse}, Mhenni et al. used drones as a case study to design a framework termed SafeSysE, which merges safety assessment and systems engineering to provide safety aspects. The objective of this paper is to develop an approach that automatically generates safety artefacts by adding safety-related concepts to the system. This approach includes a Model-Based Systems Engineering (MBSE)\cite{yakymets2013model} and a Model-Based Safety Analysis (MBSA)\cite{lisagor2011model}. This approach integrates techniques such as {Failure Mode, Effects Analysis and Fault Tree Analysis for safety analysis.} This process is not fully prototyped and has not been tested in real scenarios. In our case, we designed a safety assurance framework and applied our methodology to real-world scenarios.

The works presented in \cite{sanz2015safe} and \cite{sanz2012risk}, used ISO 31000 standards and their approach included identification, assessment and reduction procedures. The papers show how this approach has been applied in agriculture to find the sources of hazards when using UAVs in performing agricultural missions. However, the papers only provided a description of each step in the approach without any specific details on how to be validated. The experimental prototype is missing as the paper does not demonstrate a real scenario. In the current paper, we combined two standards ISO 13849 and ISO 12100 to derive a functional safety methodology for drone operations, and after that we applied this methodology {to} a real UAV use case.  
 
In \cite{barr2017preliminary}, the authors proposed two approaches, qualitative and quantitative, in performing the risk analysis process for small unmanned aircraft systems. The first  one used a safety-risk management process to identify hazards and the second approach used {a comprehensive probabilistic model based on} Bayesian network for risk estimation. {The proposed system mainly addressed some hazards without, assigning risk level for each hazard, specifying the correspondent mitigation strategies nor functional safety}, while in the current paper, we perform a detailed risk analysis of UAV controlled over the internet including some IoD-specific hazards based on international safety standards. 
In \cite{neff2016identifying}, the authors focused on the identification of human factors errors and mitigation techniques in UAV systems.
 
{Sankararaman et al. \cite{sankararaman2017towards}, developed a framework for identifying and predicting the occurrence of various risk factors that affect the safe operation of UAVs. Their work only
analyses a simple case of risk factors, i.e. battery discharging and collision while in our paper we treat various risk factors and hazards that affect the operation of UAVs.}

\section{Safety Analysis Methodology}

{Our methodology consists in using two approaches for safety analysis: (1) a qualitative functional safety analysis approach, and (2) a quantitative analysis based on the Bayesian Networks approach.}

{The first approach combines the safety standards ISO 12100 and ISO 13849. We conduct an analysis to identify the hazardous conditions, along with their possible causes and their consequences that affect the safe operation of UAVs. In addition, risk mitigation strategies required to reduce the associated risk to an acceptable level are  outlined and associated with the illustrative UAV use case, i.e. courier delivery with the drone.} 

{The second approach is a quantitative evaluation method based on Bayesian Networks used to model the safety risks related to UAV flights and to estimate the probabilities of the UAV risk of crash. A Bayesian model was designed based on the output of the hazard identification step defined in the first approach (refer to Figure \ref{fig.1}), to conduct a more detailed analysis of the relations between the risks of drone's crashes and their causes. Using this model, we assess the UAV crash probability on an illustrative scenario.}
For background information about the ISO 12100 and 13849 standards and Bayesian Networks, the reader may refer to references \cite{iso201012100}, \cite{iso200613849} and {\cite{nielsen2009bayesian}}.

In what follows, we will first present the qualitative functional safety approach, and then the quantitative Bayesian approach. The video demonstration of this methodology is available at \cite{youtube}.

\subsection{Qualitative Safety Analysis: Functional Safety approach}

\begin{figure}[htb!]
	\centering
	\includegraphics[width=0.5\textwidth ]{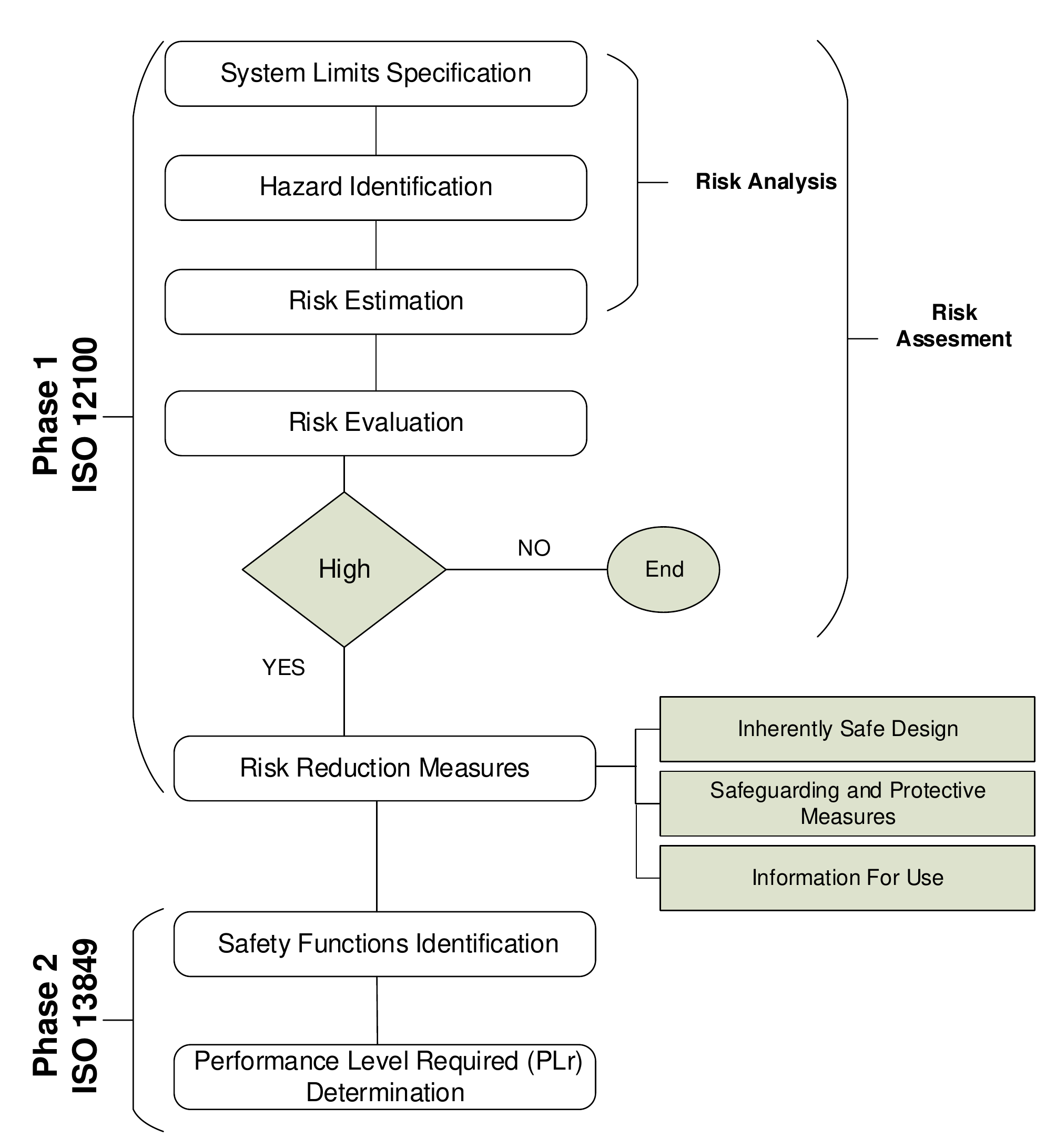}
	\caption{The Functional Safety Methodology based on ISO 12100 and ISO 13849.}
	\label{fig.1}
\end{figure}

{The proposed approach consists in a structured methodology that combines the ISO 13849 and ISO 12100 standards in order to derive the safety requirements of the drone operations when tele-operated and monitored through the Internet (see Figure \ref{fig.1}).}

\paragraph*{Description of the illustrative use case}
{To illustrate the functional safety methodology approach, a drone delivery use case is used to validate the effectiveness of the proposed approach, since the drone delivery service has become an emerging topic for different companies. As a matter of fact, Google \footnote{(2015) Google delivery drone. [Online]. Available: http://www.techspot.com/news/62412-two-delivery-drones-built-google-soontested-us.html}, DHL post service in Germany \cite{heutger2014unmanned} and Amazon in the U.S. \footnote{(2015) Amazon. [Online]. Available: http://appleinsider.com/articles/15/11/30/amazon-teases-new-details-of-planned-prime-air-drone-delivery-service.}, and many others are using drones to deliver packages to customers. }

{We consider the case of a drone that goes from a source location to a destination in an urban city, and will fly at an altitude less than 100 meters. The drone may fly on top of people, highways, streets, etc. When possible the path of the drone will be planned above non populated zones, but it might also fly {over} people to use the shortest path. The deployment conditions are important to consider when evaluating the hazards and their corresponding risks as it will {determine} the severity of the risks. In fact, flying above populated area will induce risks that are more critical than if flying on top of forests or deserts. 
We also assume that the drone communicates with the cloud using 4G connection, and the cloud relay the stream coming from the drone to a ground station/user. This architecture is similar to that proposed in \cite{KOUBAA201946, koubaa2017service}}.

{
It has to be noted that this use case is {merely} illustrative and can be applied to other uses cases of drones applications controlled over the Internet. We assume that the drone is connected to the user through the Internet, which imposes additional challenges that may lead to hazards such as network delays and message losses. We will discuss these issues in our analysis of the two approaches.}

{As shown in Figure \ref{fig.1}, the functional safety methodology has two phases: (\textit{i.}) Phase 1 is related to the standard ISO 12100, (\textit{ii.}) Phase 2 pertains to the standard ISO 13849. In what follows, we describe the functional safety analysis of the two phases for the drone delivery use case.}

\subsubsection{Phase 1: The ISO 12100 Standard}

\paragraph{\textbf{Risk Analysis:}}
{First, according to ISO 12100, a risk analysis must be carried out, which starts with \textit{system limits specification}. The drone's system limits are depicted in Table I .} {According to \cite{sanz2015safe}}, the drone limits can be divided into {five categories}: (\textit{i.}) Physical, (\textit{ii.}) Temporal, (\textit{iii.}) Environmental and (\textit{iv.}) Behavioral limits, (\textit{iv.}) Networking limits. {Table I explains the {five categories} and provides concrete examples and description of the limits. It has to be noted that the networking limits are related to the communication with the drones through the Internet, as compared with traditional line-of-sight point-to-point communication.}

\begin{table*}[htb!]
\begin {center}
\caption{Limits of the drone system.}\label{tb1}
\begin{tabular}{|l|p{14cm}|} \hline\hline
\textbf{Nature}  & \textbf{Description} \\ \hline
Physical Limits & Maximum take-off weight, maximum speed and maximum/minimum height. \\ \hline
Temporal Limits  & Maximum time of flight, response time of the commands or acquisition time of the sensors, battery degradation over time, battery life.\\ \hline
Environmental Limits  & Weather conditions (wind speed, ambient light, or dust/rain presence), the minimum distance from populated areas or from airports. \\ \hline
Behavioral Limits & Actions performed by the pilot (both autonomous and manual).\\\hline
{Networking Limits} & {Network delays, jitters, available bandwidths, latency, link availability and traffic congestion}\\\hline \hline
\end{tabular}
\end {center}
\end{table*}

The second step of the risk analysis consists in performing \textit{hazard identification}. {Table II provides the list of potential drone hazards} according to their sources, \textit{external} and \textit{internal}. This list was identified using \textit{reactive methods}. 

Reactive methods are incidents and accidents databases, safety and flight reports, survey, maintenance reports\cite{wackwitz2015safety}. 
In what follows, we provide a list of commonly used reactive methods:
\begin{itemize}
	\item \textbf{Accidents and incidents databases:}  (\textit{i.}) Federal Aviation Administration preliminary reports of Unmanned Aircraft Systems Accidents and Incidents (FAA UAS A\&I) database, (\textit{ii.}) National Transportation Safety Board (NTSB), (\textit{iii.}) FAA\textquotesingle s Aviation Safety Information Analysis and Sharing (ASIAS) database, (\textit{iv.}) FAA\textquotesingle s Accident and Incident Data System (AIDS) and Drone Crash Database \footnote{(2016) Drone Crash Database. [Online]. Available: https://dronewars.net/drone-crash-database/.}.
	
	\item \textbf{Safety and flight reports:} (\textit{i.}) NASA\textquotesingle s Aviation Safety Reporting System (ASRS), (\textit{ii.}) Annual Insurance Report \footnote{(2012) Annual Insurance Report. [Online]. Available: http://www.modelaircraft.org/files/500-q.pdf.} and (\textit{iii.}) ICAO Safety Report \footnote{(2014) ICAO Safety Report. [Online]. Available: http://www.icao.int/safety/Documents/ICAO\_2014\%20Safety\%20Report
		\_final\_02042014\_web.pdf.}.

	\item {\textbf{Maintenance reports:} FAA\textquotesingle s Aviation Maintenance Reports \footnote{(2016). Aviation Maintenance Alerts. [Online]. Available: https://www.faa.gov/aircraft/safety/alerts/aviation\_maintenance/. }.}

	\item {\textbf{Surveys:} NTSB\textquotesingle s Review of Aircraft Accident Data \footnote{ (2011) [Online]. Available: https://www.ntsb.gov/investigations/data/
		Documents/ARA1401.pdf.}, FAA\textquotesingle s Summary of Unmanned Aircraft Accident/Incident Data \footnote{ (2018) [Online]. Available: https://www.faa.gov/about/initiatives
		/maintenance\_hf/library/documents/media/human\_factors\_maintenance/a
		\_summary\_of\_unmanned\_aircraft\_accident\-incident\_data.human\_factors
		\_implications.doc.} and review of previous research papers \cite{fernando2017survey,belcastro2016aircraft,clothier2015safety,boyd2015causes,enomoto2013preliminary}.}

	\end{itemize}

\begin{table*}[htb]
\caption{Analysis of hazards sources.}\label{tb2}
\resizebox{\textwidth}{!}{%
\begin{tabular}{|c|c|l|}
\hline
\textbf{Source}                    & \textbf{Type}            & \multicolumn{1}{c|}{\textbf{Examples}}                                                                                                                                                                             \\ \hline
\multirow{7}{*}{\textbf{External}} & Interference             & \begin{tabular}[c]{@{}l@{}}{Electromagnetic interference (EMI)}, {humans (eavesdropping of radio signals)},communication interference.\end{tabular}                                                                 \\ \cline{2-3} 
                                   & Environmental conditions & {Wind, temperature, atmospheric attenuation, icing, precipitation, visibility (day or night)}.                                                                                                                       \\ \cline{2-3} 
                                   & Obstacles                & {Fixed Obstacles (trees, electric cables, buildings), and dynamic obstacles (bird, cars)}.                                                                                                                                 \\ \cline{2-3} 
                                   & Navigational Environment & \begin{tabular}[c]{@{}l@{}}GPS Signal loss/error, {GPS spoofing}, ADS-B signal inaccuracy, navigation system error, attitude error,{erroneous waypoint}.\end{tabular}                        \\ \cline{2-3} 
                                   & Air traffic environment  & Another Aircraft in close proximity, classes of airspace that may be flown nearby.                                                                                                                                 \\ \cline{2-3} 
                                   & Electrical environment   & \begin{tabular}[c]{@{}l@{}}Man-made or natural RF fields such as High Intensity Radio Transmission Areas (HIRTAs), Electrostatic phenomena.\end{tabular}                                                         \\ \cline{2-3} 
                                                                      & {Communication}   & \begin{tabular}[c]{@{}l@{}}Network Congestion, network unavailability/delays, Network Jitters.\end{tabular}                                                         \\ \cline{2-3}
                                   & Human factor             & \begin{tabular}[c]{@{}l@{}}Lack of safety culture awareness,  security attacks (on the Ground Control Station, on the DataLink, on UAV),\\ pilot error (inexperienced pilots, not familiar with the area, fatigue, rush).\end{tabular} \\ \hline
\multirow{7}{*}{\textbf{Internal}} & Mechanical               & Mechanical fastener failure, actuation failure, motor.                                                                                                                                                             \\ \cline{2-3} 
                                   & Thermal                  & Freeze, explosions.                                                                                                                                                                                                \\ \cline{2-3} 
                                   & Electronic               & Power loss, propulsion failure, saturation, overflows.                                                                                                                                                             \\ \cline{2-3} 
                                   & Algorithmic              & Verification error, decision-making error, delayed responses, infinite loops.                                                                                                                                      \\ \cline{2-3} 
                                   & Technical factor         & \begin{tabular}[c]{@{}l@{}}Battery depletion, {faulty battery cell, power loss}, inherent technical flaws (i.e., design or production),\\  technical malfunction, inappropriate charge cycle, loss of control, loss of transmission.\end{tabular}           \\ \cline{2-3} 
                                   & Software                 & \begin{tabular}[c]{@{}l@{}}Control system failure, flight control system / verification error, autopilot error, system operation error,\\  bugs in code, process errors, {vision system failure}.\end{tabular}       \\ \cline{2-3} 
                                   & Hardware                 & CPU error, avionics hardware, flight sensors.                                                                                                                                                                      \\ \hline
\end{tabular}%
}
\end{table*}

This list can be used in a straightforward manner as a checklist during hazard identification to particularly prevent inexperienced users of the drone from missing important hazards. In Table III and Table IV, we present several possible hazardous events and risk factors and their respective categories, including temporary short-time GPS loss during flight, permanent loss of GPS during flight, degraded communication quality, permanent loss of communication with ground station, security attack on the drone, loss of UAV electrical power, autopilot controller module failure, failure/inability to avoid collision, pilot error, midair collision and weather effects on UAV that affect the operation of the drone during the delivery mission.
It has to be noted that the risk assessment levels, severity levels and probabilities in Table  III and Table IV are assigned based on our personal understanding of the illustrated use case considered in this study.

For the case of a drone delivery, drones are controlled over
the Internet; hazards such as degradation of communication quality can appear
caused by networking issues such as network delays, limited
bandwidth and network congestion that affects negatively the drone performance and operation causing collision with buildings and damage to UAV.

The third step of the risk analysis is the UAV \textit{risk estimation}, which measures the underlying probabilities and severity levels of the consequences of the identified safety hazards of the drone operation. 
According to the ISO 12100 standard, the risk estimation consists in determining two parameters: (\textit{i.}) the risk severity and (\textit{ii.}) risk probability. 

The \textit{risk severity} level is estimated based on the injury level or the harmful impact on people, drone and environment. The severity of the hazard is usually affected by the consequences. 
We adopt the following four categories and their definitions of the severity levels: 
\begin{itemize}
    \item \textit{Catastrophic}: the hazard causes harm or serious injuries or deaths to humans. The severity of such hazards is the highest considering that it affects human safety and thus must be carefully addressed and removed to avoid fatal situations. For example, consider the case of the drone delivery use case where the drone has to fly on top of highways. The permanent loss of communication with the Ground Station (due to control system failure, electromagnetic interference, ...) may lead to a catastrophic situation as it results in having the drone falling on the road and thus leading to accidents. These accidents could result into injuries yet also deaths, which makes the severity \textit{catastrophic}, if no failsafe operations are implemented. 
     
    \item\textit{Critical}: the event has effect on third parties other than people, like for example making damage to building or assets in general. As an example, consider the navigation of the drone in space at relatively low altitude and the hazard that the anti-collision sensors stop functioning. This hazard can lead to a collision and a crash against a building and may lead to damages to assets if it falls on the ground, or breaking some window glasses of the building, and thus is tagged of a \textit{critical} severity. 
    \item\textit{Marginal}: the event causes damages to the drone system itself. For example, when the drone is landing in a certain open space, it is possible to lose control due to interference to IMU and altitude sensors or GPS signals, which may lead to the case of crashing against the ground. This hazard will only lead to crashing the drone or some of its part, with no damages on third parties or people. Then, its severity is tagged as \textit{marginal}. 
    \item\textit{Negligible}: { the event does not affect the operational capability of the drone but causes minor effects on drone system performance (mission degradation). For example, the number of GPS satellites may decrease in some areas with high building which may affect the accuracy of localization of the drone in space. This hazard does not lead to crashes by usually it leads to temporary high localization error that remain acceptable.  This hazard is tagged as \textit{negligible}}.
\end{itemize}

{It has to be noted} that the \textit{risk severity} is particularly susceptible to bias or incorrect assumptions, as severity is generally subject to a qualitative analysis of an event and open to interpretations \cite{wolf2017drones}. In fact, the severity may also depend on the context. For example, we have tagged the severity level of {the degraded communication quality as \textit{critical}}, given the context it is operated on top of highways. Changing the navigation context of the application to being on top of forest or areas not having people, the severity may be tagged in this case to \textit{marginal} as it will lead only to drone crashing. Thus, when applying the methodology, the severity has to consider all playing factors of the use case and deployment assumptions to be accurately interpreted.

{On the other hand, the \textit{risk probability} is defined as the likelihood that the consequence of the safety hazard might occur. }

{As the analysis is qualitative, the probabilities are not expressed as numerical values, but as attributes. According to \cite{prob}, the range of probabilities of occurrence is divided into five main classes:}
\begin{itemize}
    \item {\textit{Frequent}: the event is likely to occur many times or has occurred frequently, }
    \item {\textit{Probable}: likely to occur regularly but not frequent, }
    \item {\textit{Occasional}: likely to occur sometimes or has occurred infrequently, }
    \item {\textit{Remote}: unlikely to occur but possible or has occurred rarely, }
    \item {\textit{Improbable}: very unlikely to occur or not known to have occurred.}
\end{itemize}

\begin{figure}[htb!]
\begin{center}
\includegraphics[width=0.5\textwidth]{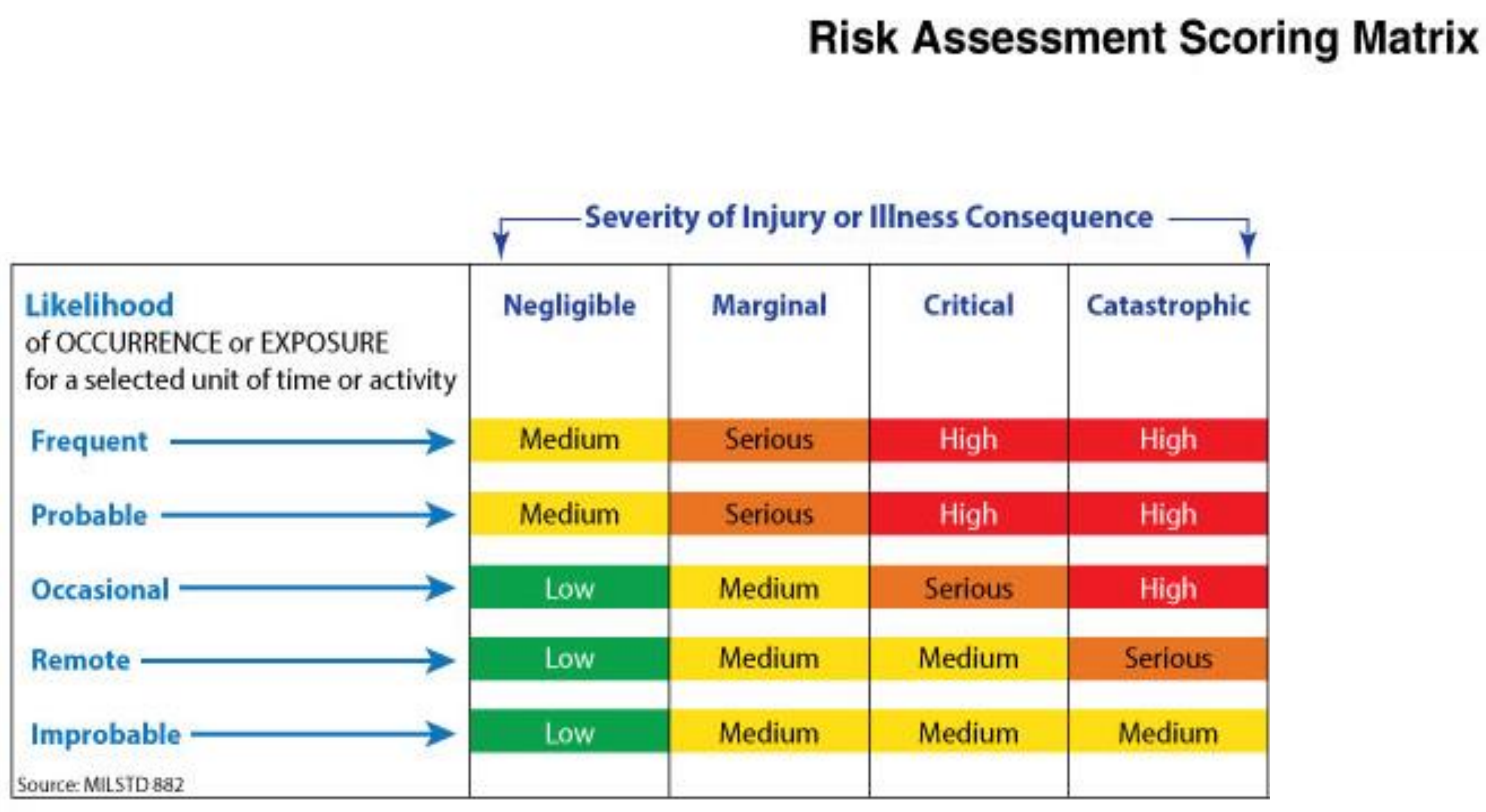}
\caption{A typical risk assessment scoring matrix (based on ISO 12100 Safety of Machinery)}
\label{fig.2}
\end{center}
\end{figure}

{As a first example, we consider the hazard of a permanent loss of GPS signals during flight, which may result in a drone losing direction and crashes (see Table III and Table IV). A GPS loss could potentially result in a collision with an aircraft, UAS, or a damage to UAV or environment and ground, and even humans, thus severity is \textit{catastrophic}. In the condition of open air environment and hovering at high altitude, losing GPS signals could be seen as \textit{remote}. Using the risk assessment matrix of Figure \ref{fig.2}, the risk of GPS loss is thus assessed as "\textit{serious}". }
{\footnotetext{A Matter of Risk Assessment, Liability \& Compliance: Machine Safety Labeling in the 21st Century. Accessed: JULY 23, 2014. [Online]. Available:
https://www.automation.com/automation-news/article/a-matter-of-risk-assessment-liability-compliance-machine-safety-labeling-in-the-21st-century.
}}

{As a second example, the degradation of the communication quality is another hazard related to the context of the Internet-of-Drones. The degradation of QoS may be due to several reasons such as network congestion, delays, jitters, communication instability, etc. Assuming to use a 4G connection, the communication quality is most often stable and reliable, but be subject to perturbation in areas with bad coverage or due to signal attenuation. Therefore, this hazard can be seen to occur \textit{occasionally} for a typical 4G connection. {The degradation of the communication quality may result in critical consequences (i.e. Severity is \textit{critical}) as it results in intermediate interruption of the monitoring during an autonomous mission and can lead to collision with buildings.}  Using the risk assessment matrix of Figure \ref{fig.2}, the risk caused by degradation of communication quality is estimated to {"\textit{serious}"}. Of course, using another type of communication that is less reliable, would lead to a different risk estimation. }
\afterpage{%
    \clearpage
    \thispagestyle{empty}
    \begin{landscape}
\begin{table}[htb!]

\caption{Application of Functional Safety Methodology {for Drone Delivery Application}}
\label{my-label}
\centering
\resizebox{!}{.295\paperheight}{%

\renewcommand{\arraystretch}{2}
\begin{tabular}{|l|l|l|l|l|l|l|l|l|l|l|l|l|l|l|l|l|l|}
\hline
\multicolumn{14}{|c|}{\textbf{ISO 12100}} & \multicolumn{4}{c|}{\textbf{ISO 13849}} \\ \hline
\multicolumn{7}{|c|}{Hazard Identification} & \multicolumn{3}{c|}{Risk Assessment} & \multicolumn{4}{c|}{Risk Reduction} & \multicolumn{4}{c|}{\begin{tabular}[c]{@{}c@{}}Performance Level \\ Required Determination\end{tabular}} \\ \hline
\multicolumn{1}{|c|}{ID} & \multicolumn{1}{c|}{Hazard} & \multicolumn{1}{c|}{Source} & \multicolumn{1}{c|}{Type} & \multicolumn{1}{c|}{Element} & \multicolumn{1}{c|}{Cause} & \multicolumn{1}{c|}{Consequence} & \multicolumn{1}{c|}{P} & \multicolumn{1}{c|}{S} & \multicolumn{1}{c|}{Risk Level} & \multicolumn{1}{c|}{\begin{tabular}[c]{@{}c@{}}Risk Reduction\\  Measures\end{tabular}} & \multicolumn{1}{c|}{\begin{tabular}[c]{@{}c@{}}By Inherently\\  Safe Design\end{tabular}} & \multicolumn{1}{c|}{\begin{tabular}[c]{@{}c@{}}By \\ Safeguarding\end{tabular}} & \multicolumn{1}{c|}{\begin{tabular}[c]{@{}c@{}}By Information \\ For Use\end{tabular}} & \multicolumn{1}{c|}{S} & \multicolumn{1}{c|}{F} & \multicolumn{1}{c|}{P} & \multicolumn{1}{c|}{PLr} \\ \hline
\multirow{2}{*}{1} & \multirow{2}{*}{\begin{tabular}[c]{@{}l@{}}Temporary Short-Time\\ GPS Loss during flight\end{tabular}} & \multirow{2}{*}{External} & \multirow{2}{*}{Interference} & \multirow{2}{*}{UAV} & \multirow{2}{*}{\begin{tabular}[c]{@{}l@{}}Interference, going through a tunnel,\\ going through high buildings, ...\end{tabular}} & \multirow{2}{*}{\begin{tabular}[c]{@{}l@{}}Temporary loss of navigation control,\\ unstable UAV,\\ small deviation from planned path\end{tabular}} & \multirow{2}{*}{Probable} & \multirow{2}{*}{Negligible} & \multirow{2}{*}{Medium} & \begin{tabular}[c]{@{}l@{}}Use high-quality \\ GPS devices (e.g. RTK)\end{tabular} & x &  &  & \multicolumn{4}{l|}{} \\ \cline{11-18} 
 &  &  &  &  &  &  &  &  &  & \begin{tabular}[c]{@{}l@{}}Check for radio interference that\\ may compromise communication\\ signals\end{tabular} & x &  &  & \multicolumn{4}{l|}{} \\ \hline
\multirow{5}{*}{2} & \multirow{5}{*}{\begin{tabular}[c]{@{}l@{}}Permanent Loss of GPS\\ during flight\end{tabular}} & \multirow{5}{*}{External} & \multirow{5}{*}{Electronic} & \multirow{5}{*}{UAV} & \multirow{5}{*}{\begin{tabular}[c]{@{}l@{}}Defect of GPS device,\\ loss of GPS signal\end{tabular}} & \multirow{5}{*}{\begin{tabular}[c]{@{}l@{}}Control loss, collision with UAS,\\ crashing to the ground, UAVs\\ can lead to injuries to people if\\ flight is on top of populated zone,\\ UAV damage\end{tabular}} & \multirow{5}{*}{Remote} & \multirow{5}{*}{Catastrophic} & \multirow{5}{*}{Serious} & \multirow{4}{*}{\begin{tabular}[c]{@{}l@{}}Use high-quality\\ GPS devices (e.g. RTK)\end{tabular}} & \multirow{4}{*}{x} & \multirow{4}{*}{} & \multirow{4}{*}{} & \multicolumn{4}{l|}{\multirow{4}{*}{}} \\
 &  &  &  &  &  &  &  &  &  &  &  &  &  & \multicolumn{4}{l|}{} \\
 &  &  &  &  &  &  &  &  &  &  &  &  &  & \multicolumn{4}{l|}{} \\
 &  &  &  &  &  &  &  &  &  &  &  &  &  & \multicolumn{4}{l|}{} \\ \cline{11-18} 
 &  &  &  &  &  &  &  &  &  & \begin{tabular}[c]{@{}l@{}}Minimize navigation paths\\ over urban areas and highways\end{tabular} & x &  &  & \multicolumn{4}{l|}{} \\ \hline
\multirow{5}{*}{3} & \multirow{5}{*}{\begin{tabular}[c]{@{}l@{}}Degraded Communication\\ Quality\end{tabular}} & \multirow{5}{*}{External} & \multirow{5}{*}{Communication} & \multirow{5}{*}{UAV, GCS, Cloud} & \multirow{5}{*}{\begin{tabular}[c]{@{}l@{}}4G unstable Connection,\\ Network Congestion,\\ Long Delays,\\ Network Jitters, ...\end{tabular}} & \multirow{5}{*}{\begin{tabular}[c]{@{}l@{}}Temporary loss of monitor and control,\\ occasional command losses,\\ collision with buildings,\\ damage to UAV\end{tabular}} & \multirow{5}{*}{Occasional} & \multirow{5}{*}{Critical} & \multirow{5}{*}{Serious} & \begin{tabular}[c]{@{}l@{}}Use a network with a\\ guaranteed quality of service\end{tabular} & x &  &  & \multicolumn{4}{l|}{} \\ \cline{11-18} 
 &  &  &  &  &  &  &  &  &  & \begin{tabular}[c]{@{}l@{}}Implement failsafe mechanisms\\ when connection is lost\end{tabular} &  & x &  & S1 & F2 & P2 & c \\ \cline{11-18} 
 &  &  &  &  &  &  &  &  &  & \begin{tabular}[c]{@{}l@{}}Verification and prototyping\\ through extensive network simulations\end{tabular} & x &  &  & \multicolumn{4}{l|}{} \\ \cline{11-18} 
 &  &  &  &  &  &  &  &  &  & \begin{tabular}[c]{@{}l@{}}Monitor the communication\\ quality in real time\end{tabular} &  &  & x & \multicolumn{4}{l|}{} \\ \cline{11-18} 
 &  &  &  &  &  &  &  &  &  & Log files &  &  & x & \multicolumn{4}{l|}{} \\ \hline
\multirow{5}{*}{4} & \multirow{5}{*}{\begin{tabular}[c]{@{}l@{}}Permanent Loss of Communication\\ with Ground Station\end{tabular}} & \multirow{5}{*}{External} & \multirow{5}{*}{Communication} & \multirow{5}{*}{UAV, GCS, Cloud} & \multirow{5}{*}{\begin{tabular}[c]{@{}l@{}}Control system failure,\\ environmental condition,\\ power loss, software verification\\ error and EMI.\end{tabular}} & \multirow{5}{*}{\begin{tabular}[c]{@{}l@{}}Crash into building, obstacle,\\ injuries to people, vehicle damage,\\ undesired flight trajectory,\\ uncontrolled maneuvers, loss of vehicle control\end{tabular}} & \multirow{5}{*}{Probable} & \multirow{5}{*}{Catastrophic} & \multirow{5}{*}{High} & \multirow{4}{*}{Preflight inspection} & \multirow{4}{*}{x} & \multirow{4}{*}{} & \multirow{4}{*}{} & \multicolumn{4}{l|}{\multirow{4}{*}{}} \\
 &  &  &  &  &  &  &  &  &  &  &  &  &  & \multicolumn{4}{l|}{} \\
 &  &  &  &  &  &  &  &  &  &  &  &  &  & \multicolumn{4}{l|}{} \\
 &  &  &  &  &  &  &  &  &  &  &  &  &  & \multicolumn{4}{l|}{} \\ \cline{11-18} 
 &  &  &  &  &  &  &  &  &  & \begin{tabular}[c]{@{}l@{}}Monitoring the data link\\ performance during the mission\end{tabular} &  &  & x & \multicolumn{4}{l|}{} \\ \hline
\multirow{4}{*}{5} & \multirow{4}{*}{Security attack on the drone} & \multirow{4}{*}{External} & \multirow{4}{*}{Software} & \multirow{4}{*}{UAV, GCS, Cloud} & \multirow{4}{*}{Communication protocol insecure (e.g. MAVLink)} & \multirow{4}{*}{\begin{tabular}[c]{@{}l@{}}Drone control loss,\\ criminal attacks using the drone,\\ drone hijack\end{tabular}} & \multirow{4}{*}{Probable} & \multirow{4}{*}{Critical} & \multirow{4}{*}{High} & \multirow{3}{*}{\begin{tabular}[c]{@{}l@{}}Secure the communication protocols\\ between the drone, cloud and ground station\end{tabular}} & \multirow{3}{*}{x} & \multirow{3}{*}{} & \multirow{3}{*}{} & \multicolumn{4}{l|}{\multirow{3}{*}{}} \\
 &  &  &  &  &  &  &  &  &  &  &  &  &  & \multicolumn{4}{l|}{} \\
 &  &  &  &  &  &  &  &  &  &  &  &  &  & \multicolumn{4}{l|}{} \\ \cline{11-18} 
 &  &  &  &  &  &  &  &  &  & failsafe mechanism &  & x &  & S1 & F2 & P2 & c \\ \hline
\multirow{6}{*}{6} & \multirow{6}{*}{Loss of UAV electrical power} & \multirow{6}{*}{Internal} & \multirow{6}{*}{Technical} & \multirow{6}{*}{UAV} & \multirow{6}{*}{\begin{tabular}[c]{@{}l@{}}Faulty battery cell, faulty charge,\\ inappropriate charge cycle,\\ manufacturing defect, vibration\end{tabular}} & \multirow{6}{*}{\begin{tabular}[c]{@{}l@{}}Degraded flight,\\ harm to people, crash\end{tabular}} & \multirow{6}{*}{Occasional} & \multirow{6}{*}{Catastrophic} & \multirow{6}{*}{High} & \begin{tabular}[c]{@{}l@{}}Check of battery in\\ preflight phase\end{tabular} & x &  &  & \multicolumn{4}{l|}{} \\ \cline{11-18} 
 &  &  &  &  &  &  &  &  &  & \begin{tabular}[c]{@{}l@{}}Verifying the design\\ content of the drone\end{tabular} & x &  &  & \multicolumn{4}{l|}{} \\ \cline{11-18} 
 &  &  &  &  &  &  &  &  &  & Warning system &  &  & x & \multicolumn{4}{l|}{} \\ \cline{11-18} 
 &  &  &  &  &  &  &  &  &  & Parachute &  & x &  & S2 & F1 & P2 & d \\ \cline{11-18} 
 &  &  &  &  &  &  &  &  &  & \begin{tabular}[c]{@{}l@{}}Flying above non\\ populated areas\end{tabular} &  &  & x & \multicolumn{4}{l|}{} \\ \cline{11-18} 
 &  &  &  &  &  &  &  &  &  & Real time battery information &  &  & x & \multicolumn{4}{l|}{} \\ \hline
\end{tabular}%
}

\end{table}
\end{landscape}
    \clearpage
}

The result of the risk assessment phase consists of a prioritized list of hazards and their corresponding risk levels, as shown in Table III and Table IV. 

\paragraph{\textbf{Risk Evaluation:}}
The risk evaluation step is defined based on the results of the risk estimation step. In this evaluation step, we identify all the risks that are not tolerable and then we will process them in the risk reduction step. If all the risks are acceptable, then, there is no need to proceed further (refer to Figure \ref{fig.1}).

\paragraph{\textbf{Risk Reduction:}}
{After the evaluation of risks, the next step deals with the risk reduction. The process consists in identifying the hazards that led to an unacceptable level of harm; then, try to reduce the risk to a tolerable level.}

{The risk reduction measures for the drone delivery use case are described below. According to ISO 12100, there are three steps to follow for risk mitigation: (1) inherently safe design measures, (2) safeguarding and (3) information for use \cite{jespen2016risk}}.

{In the first step, the strategy consists in following safe design approaches without the use of safeguards or protective measures to reduce the risk to an acceptable level in the design phase, and to ensure the safety of the system. These approaches include: }

\begin{itemize}
		\item { Perform a safe design of the drone in order to improve its stability, to avoid causing harms to a person.}
		\item {Prototyping and verification.}
		\item {Visual inspection for verifying and checking the design content in case problems arise with the drone before flying.}
\end{itemize}

{ Applying the above principles in our context; in the case of Internet-of-Drones, where communication occurs through the Internet, it is important to} { use a designed network with a guaranteed quality of service in terms of delay and throughput. 
 
Then, the network design must be verified to ensure that it operates as expected. The verification and prototyping can be done through extensive network simulations. Furthermore, as part of the visual inspection, it is important to constantly monitor the quality of service of the communication in real-time to avoid any possible hazard resulting from bad communication.
}

{In the second step, it is recommended to take possible safeguards and (technical) protective measures that help to mitigate remaining risks. In ISO 12100, the safeguards are defined as "\textit{protective measures to protect persons from the hazards which cannot reasonably be eliminated or risks which cannot be sufficiently reduced by inherently safe design measures}" \cite{iso201012100}. For example, for the case of midair collision risk, we can use parachute, airbags, or protection nets as possible safeguards and protective measures. Collision avoidance sensors and propeller guards can be used to avoid collisions to walls.}

{ In the case of 4G communication, it is possible to implement failsafe mechanisms to prevent from risk of crashes,
reduce any further damage of UAV and ensure the safe drone operation, for example, the drone should land or return to home in case of loss of wireless communication.  }

{
In the third step, information for uses are necessary
to reduce risk and ensures the safe drone operation. According
to ISO 12100, "\textit{information for uses are protective measures
consisting of lists of elements of information (for example,
text, words, instructions, warning signs, markings and labels,
audible or visual signals) used to convey information to the
user, which may be essential for keeping risks on an acceptable low risk level"} \cite{iso201012100}. In the case of 4G communication, it is possible to implement information for use measures, which are classified into three categories:}

\begin{itemize}
	\item {\texttt{Preflight:} Before the flight, it is important to make sure that the communication quality is up to a certain acceptable level before starting the mission. In case of bad communication is detected through sensor readings signals , the drone will be prevented from taking-off and starting its mission until the communication is resumed at a good level.} 
	\end{itemize}

\begin{itemize}
	\item {\texttt {During the mission:}	The communication quality should be monitored in real-time and in case of a high degradation, the drone should operate autonomously in a safe manner without depending on the control of the ground station/user. }
\end{itemize}
\begin{itemize}
	\item {\texttt {Post-flight:} After the mission is completed,  the log files can be analyzed to understand the reasons behind the occurrence of risks during the mission (if any) and develop additional protective measures to avoid bad consequences when similar risks occur in the future. }
	 
\end{itemize}

{In the case of loss of UAV electrical power, warning system for alerting UAV pilot of battery failure are required and real time battery information can be useful to take necessary actions to avoid the consequence of this hazard. }

{In Table III and Table IV, we present all measures that might be suitable for reducing the risks of the identified hazardous events. We remind that the risk assessments, the severity levels and probabilities are detected based on our understanding of the illustrative use case of drone delivery. }

{After completion of risk assessment and risk reduction following ISO 12100, the result of risk reduction is required to implement protective measures (safeguards) employing safety functions (are often called safety-related parts of control systems (SRP/CS)) in order to eliminate hazard and/or reduce risk. This will lead to part of ISO 13849 (Phase 2) to provide safety specifications and guidance on the determination of a required (adequate) level of Functional Safety in the form of a required Performance Level (PLr).}

\subsubsection{Phase 2: The ISO 13849 Standard}

\paragraph{\textbf{Safety Functions Identification:}} 

{The first step of ISO 13849 is to identify the safety functions to be performed by SRP/CS. The safety functions can be used as \textit{safeguarding and protective measures}.
The best way is to take this information directly from the results of risk reduction measures performed in the ISO 12100 phase (Refer to Table III and Table IV).}

\paragraph{\textbf{Performance Level Required Specification:}}

{Once the safety functions are identified, the required performance level (PLr) must be calculated to specify  the ability of safety-related parts of the system to perform a safety function under foreseeable conditions \cite {iso200613849}.}

{In other words, the PLr refers to the performance level that must be applied to attain the required risk reduction for each safety function. The required performance levels depend on the expected risk that originates from each hazard and has to be determined based on the risk estimation.}
\afterpage{%
    \clearpage
    \thispagestyle{empty}
    \begin{landscape}
\begin{table}[htb!]

\caption{Application of Functional Safety Methodology {for Drone Delivery Application}}
\label{my-label}
\centering
\resizebox{!}{.295\paperheight}{%

\renewcommand{\arraystretch}{2}
\begin{tabular}{|l|l|l|l|l|l|l|l|l|l|l|l|l|l|l|l|l|l|}
\hline
\multicolumn{14}{|c|}{\textbf{ISO 12100}} & \multicolumn{4}{c|}{\textbf{ISO 13849}} \\ \hline
\multicolumn{7}{|c|}{Hazard Identification} & \multicolumn{3}{c|}{Risk Assessment} & \multicolumn{4}{c|}{Risk Reduction} & \multicolumn{4}{c|}{\begin{tabular}[c]{@{}c@{}}Performance Level \\ Required Determination\end{tabular}} \\ \hline
\multicolumn{1}{|c|}{ID} & \multicolumn{1}{c|}{Hazard} & \multicolumn{1}{c|}{Source} & \multicolumn{1}{c|}{Type} & \multicolumn{1}{c|}{Element} & \multicolumn{1}{c|}{Cause} & \multicolumn{1}{c|}{Consequence} & \multicolumn{1}{c|}{P} & \multicolumn{1}{c|}{S} & \multicolumn{1}{c|}{Risk Level} & \multicolumn{1}{c|}{\begin{tabular}[c]{@{}c@{}}Risk Reduction\\  Measures\end{tabular}} & \multicolumn{1}{c|}{\begin{tabular}[c]{@{}c@{}}By Inherently\\  Safe Design\end{tabular}} & \multicolumn{1}{c|}{\begin{tabular}[c]{@{}c@{}}By \\ Safeguarding\end{tabular}} & \multicolumn{1}{c|}{\begin{tabular}[c]{@{}c@{}}By Information \\ For Use\end{tabular}} & \multicolumn{1}{c|}{S} & \multicolumn{1}{c|}{F} & \multicolumn{1}{c|}{P} & \multicolumn{1}{c|}{PLr} \\ \hline
\multirow{5}{*}{7} & \multirow{5}{*}{\begin{tabular}[c]{@{}l@{}}Autopilot controller\\ module failure\end{tabular}} & \multirow{5}{*}{Internal} & \multirow{5}{*}{Software} & \multirow{5}{*}{UAV} & \multirow{5}{*}{\begin{tabular}[c]{@{}l@{}}Timing errors, memory corruption,\\ incorrect specification,\\ incorrect implementation,\\ inaccurate/ incorrect assumptions.\end{tabular}} & \multirow{5}{*}{Loss of flight} & \multirow{5}{*}{Remote} & \multirow{5}{*}{Negligible} & \multirow{5}{*}{Low} & \multirow{5}{*}{\begin{tabular}[c]{@{}l@{}}failsafe autopilot\\ intervenes when\\ failure of autopilot\\ detected\end{tabular}} & \multirow{5}{*}{} & \multirow{5}{*}{x} & \multirow{5}{*}{} & \multirow{5}{*}{S1} & \multirow{5}{*}{F1} & \multirow{5}{*}{P1} & \multirow{5}{*}{a} \\
 &  &  &  &  &  &  &  &  &  &  &  &  &  &  &  &  &  \\
 &  &  &  &  &  &  &  &  &  &  &  &  &  &  &  &  &  \\
 &  &  &  &  &  &  &  &  &  &  &  &  &  &  &  &  &  \\
 &  &  &  &  &  &  &  &  &  &  &  &  &  &  &  &  &  \\ \hline
\multirow{9}{*}{8} & \multirow{9}{*}{\begin{tabular}[c]{@{}l@{}}Failure/Inability to\\ avoid collision\end{tabular}} & \multirow{9}{*}{External} & \multirow{9}{*}{Obstacles} & \multirow{9}{*}{UAV} & \multirow{9}{*}{\begin{tabular}[c]{@{}l@{}}Vision system failure,\\ erroneous waypoint that create\\ collision with obstacles,\\ harsh environmental condition,\\ inaccurate GPS signal,\\ inadequate design of collision\\ avoidance system,\\ another aircraft in close\\ proximity, fixed obstacles,\\ conflict with moving obstacles\end{tabular}} & \multirow{9}{*}{\begin{tabular}[c]{@{}l@{}}Collision with obstacles,\\ UAVs, vehicle damage\end{tabular}} & \multirow{9}{*}{Probable} & \multirow{9}{*}{Critical} & \multirow{9}{*}{High} & \multirow{4}{*}{\begin{tabular}[c]{@{}l@{}}Parachute, airbags,\\ propeller guards\end{tabular}} & \multirow{4}{*}{} & \multirow{4}{*}{x} & \multirow{4}{*}{} & \multirow{4}{*}{S2} & \multirow{4}{*}{F2} & \multirow{4}{*}{P1} & \multirow{4}{*}{d} \\
 &  &  &  &  &  &  &  &  &  &  &  &  &  &  &  &  &  \\
 &  &  &  &  &  &  &  &  &  &  &  &  &  &  &  &  &  \\
 &  &  &  &  &  &  &  &  &  &  &  &  &  &  &  &  &  \\ \cline{11-18} 
 &  &  &  &  &  &  &  &  &  & \begin{tabular}[c]{@{}l@{}}Re-design of Collision\\ Avoidance System\end{tabular} & x &  &  & \multicolumn{4}{l|}{} \\ \cline{11-18} 
 &  &  &  &  &  &  &  &  &  & \multirow{4}{*}{\begin{tabular}[c]{@{}l@{}}Wind speed\\ monitoring\end{tabular}} & \multirow{4}{*}{} & \multirow{4}{*}{} & \multirow{4}{*}{x} & \multicolumn{4}{l|}{\multirow{4}{*}{}} \\
 &  &  &  &  &  &  &  &  &  &  &  &  &  & \multicolumn{4}{l|}{} \\
 &  &  &  &  &  &  &  &  &  &  &  &  &  & \multicolumn{4}{l|}{} \\
 &  &  &  &  &  &  &  &  &  &  &  &  &  & \multicolumn{4}{l|}{} \\ \hline
\multirow{5}{*}{9} & \multirow{5}{*}{Pilot error} & \multirow{5}{*}{External} & \multirow{5}{*}{Human factor} & \multirow{5}{*}{UAV} & \multirow{5}{*}{\begin{tabular}[c]{@{}l@{}}Pilot not familiar with the area,\\ pilot unfamiliar with equipment,\\ inexperienced pilots,\\ lack of training\end{tabular}} & \multirow{5}{*}{\begin{tabular}[c]{@{}l@{}}Harm to people,\\ damage of drone;\\ damage of infrastructure\\ (high costs)\end{tabular}} & \multirow{5}{*}{Remote} & \multirow{5}{*}{Catastrophic} & \multirow{5}{*}{Serious} & \multirow{5}{*}{\begin{tabular}[c]{@{}l@{}}failsafe\\ mechanism\end{tabular}} & \multirow{5}{*}{} & \multirow{5}{*}{x} & \multirow{5}{*}{} & \multirow{5}{*}{S2} & \multirow{5}{*}{F1} & \multirow{5}{*}{P2} & \multirow{5}{*}{d} \\
 &  &  &  &  &  &  &  &  &  &  &  &  &  &  &  &  &  \\
 &  &  &  &  &  &  &  &  &  &  &  &  &  &  &  &  &  \\
 &  &  &  &  &  &  &  &  &  &  &  &  &  &  &  &  &  \\
 &  &  &  &  &  &  &  &  &  &  &  &  &  &  &  &  &  \\ \hline
\multirow{5}{*}{10} & \multirow{5}{*}{Midair collision} & \multirow{5}{*}{External} & \multirow{5}{*}{Air traffic environment} & \multirow{5}{*}{UAV} & \multirow{5}{*}{Vision system failure} & \multirow{5}{*}{\begin{tabular}[c]{@{}l@{}}Damage or loss of one or\\ both of the aircraft,\\ damage to property\\ and people injury.\end{tabular}} & \multirow{5}{*}{Remote} & \multirow{5}{*}{Catastrophic} & \multirow{5}{*}{Serious} & \multirow{4}{*}{\begin{tabular}[c]{@{}l@{}}Parachute, airbags,\\ or protection nets\end{tabular}} & \multirow{4}{*}{} & \multirow{4}{*}{x} & \multirow{4}{*}{} & \multirow{4}{*}{S2} & \multirow{4}{*}{F1} & \multirow{4}{*}{P2} & \multirow{4}{*}{d} \\
 &  &  &  &  &  &  &  &  &  &  &  &  &  &  &  &  &  \\
 &  &  &  &  &  &  &  &  &  &  &  &  &  &  &  &  &  \\
 &  &  &  &  &  &  &  &  &  &  &  &  &  &  &  &  &  \\ \cline{11-18} 
 &  &  &  &  &  &  &  &  &  & \begin{tabular}[c]{@{}l@{}}Collision Avoidance\\ Sensors\end{tabular} &  & x &  & S2 & F1 & P2 & d \\ \hline
11 & Weather effects on UAV & External & Environmental conditions & UAV & \begin{tabular}[c]{@{}l@{}}Severe weather or climatic events\\ (hurricanes, tornadoes,\\ thunderstorms,lightning,\\ wind shear, icing)\end{tabular} & Vehicle loss of control & Occasional & Negligible & Low & \begin{tabular}[c]{@{}l@{}}Wind speed\\ monitoring\end{tabular} &  &  & x & \multicolumn{4}{l|}{} \\ \hline
\end{tabular}
}

\end{table}
\end{landscape}
    \clearpage
}

\begin{figure}[htb!]
\begin{center}
\includegraphics[width=0.5\textwidth]{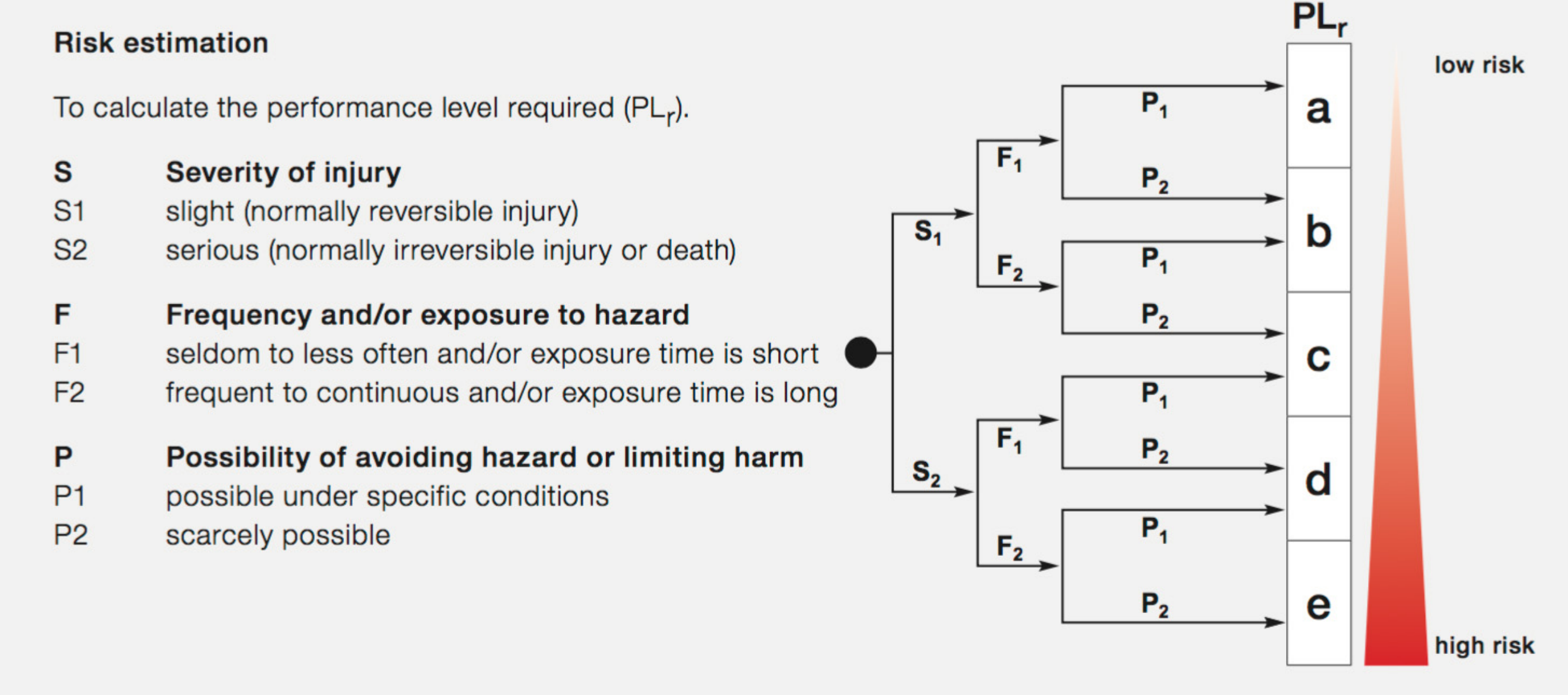}
\caption{ Risk Graph for Determining the Required Performance Level for a
Safety Function from ISO 13849. }
\label{fig.3}
\end{center}
\end{figure}

{According to ISO 13849, the required performance level PLr is estimated using a risk graph, as presented in { Fig. \ref{fig.3}}. 

The PLr is determined by estimating the three parameters\cite{iso200613849}: (\textit{i.}) severity of possible injury (S), (\textit{ii.}) frequency of exposure to hazard (F) and (\textit{iii.}) the possibility of avoiding the hazard (P), for each safety function:}
\begin{itemize}
    \item {\textit{Severity of possible injury (S)}: If the severity of injury is high or induce death, S2 is selected, but S1 is selected if the severity of injury is low.}
    \item {\textit{Frequency of exposure to hazard (F)}: Reflect the drone's degree of frequency and/or exposure to the hazard. F2 should be selected if the drone is frequently or continuously exposed to the
hazard, but F1 is selected if the time of exposure to the hazard is short.}
 \item {\textit{Possibility of avoiding the hazard (P)}: when a hazardous situation occurs, P1 should only be selected if there is a realistic chance of avoiding an accident or of significantly reducing its effect. P2 should be selected if there is almost no chance of avoiding the hazard.}
\end{itemize}

{The final output of the risk graph in Fig.\ref{fig.3} will indicate a performance level required PLr, which is used to denote what performance level is required by the safety function, which is graded ``a''to``e''. Clearly, the greater the risk of exposure to a hazard, the higher the performance of the safety related control needs to be.}

{Let us consider a case of wireless communication that is not reliable to illustrate the process on how to estimate the PLr.
In the case of a degraded communication quality due to networking issues, failsafe mechanism is a safety function that must be designed to reduce the risk related to networking issues. The degradation of communication quality can cause damage to UAV which correspond to severity level S1. A degraded communication quality could frequently be expected to occur which make the drone frequently exposed to this hazard (F2). When this hazard occurs, there is no possibility to avoid it (P2).
According to the risk graph in Fig.\ref{fig.3}, the PLr required by the failsafe mechanism for eliminating hazardous situation is ``c''.  
  } 
{\footnotetext{What are functional safety standards for servo drives?. Accessed: JANUARY 18, 2017. [Online]. Available:
https://www.motioncontroltips.com/faq\-what\-are\-functional\-safety\-standards\-for\-servo\-drives/.].
}}
  
 {One of the most dangerous hazards to pilots is the mid-air collision, which may occur remotely between two UAS systems (frequency is F1). Depending on the nature of the collision, they can result in the loss of one or both of the aircraft. A secondary accident usually following mid-air collisions is ground impact, that may injure people and damage property. Potential damages resulting from all these accidents include injury or fatality of people on the ground or on-board another aircraft, damage or loss of the vehicle and damage to property which correspond to severity level S2. When this hazard occurs, there is no possibility to avoid it (P2). According to the risk graph in Fig.\ref{fig.3}, the PLr required by the safeguards (Parachute, airbags, or protection nets and collision avoidance sensors) for eliminating hazardous situation is ``d''.}

Finally, the functional safety system must be documented. Table III and Table IV contains all information carried out in the safety risk assessment process to derive a functional safety system.

{Using these two standards, risk analysis is made in a qualitative manner. However, these standards also demand a quantitative analysis of risk. Bayesian networks can be used to quantify and improve the qualitative risk assessment.}

\subsection{Quantitative Safety Analysis: Bayesian Networks approach}

{As mentioned earlier, the qualitative method based on the functional safety assessment does not give a comprehensive analysis and must be complemented with a quantitative analysis. 
In this section, we propose a quantitative risk analysis method based on a Bayesian Networks approach to conduct a more detailed analysis of the relationships among risk of drone crashes and their causes to prevent the crash from occurring.}
First, based on the safety analysis output described in ISO 12100 and ISO 13849, the Bayes model of the UAV crash risk is created by causality; data were populated by collecting and reviewing civil UAV accidents and the expected probability of occurrence of a crash associated with UAV system deduced.
Second, a typical example of causal and diagnostic inference was demonstrated. Third, a scenario analysis was established for demonstration purposes. Finally, the result of the sensitivity analysis is discussed.

\subsubsection{The Bayesian Networks topology}

The BN model is typically composed of target, observable and intermediate nodes \cite{brito2016bayesian}. Target nodes are nodes that represent variables for which a probability distribution is computed (Probability of drone crash).

Observable nodes represent variables that are measurable or directly observable. These nodes provide the information necessary to compute the prior probability of events connected.
Intermediate nodes are mainly defined to help manage the size of the conditional probability tables (CPT), because too many parent nodes with their states result in massive CPT structures that are difficult to visualize. So, combining the parent nodes into fewer intermediate nodes based on causal structures is a practical solution. Following these instructions, we establish causal BN model between target, observable and intermediate nodes to predict the risk of a crash of UAV during a mission in different conditions.

\begin{figure*}[htb]
	\centering
		\includegraphics[width=7in,height=3in] {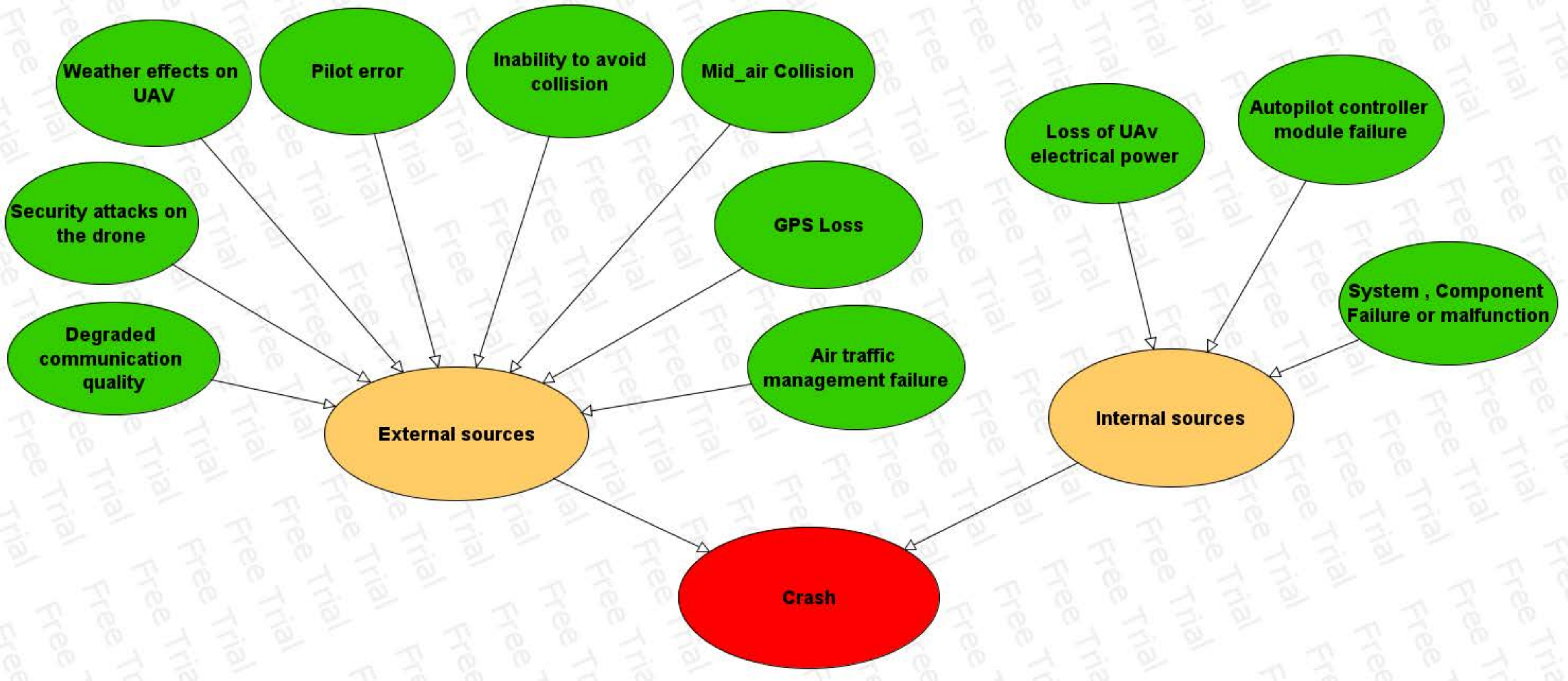}
	\caption{ A causal model for predicting UAV risk of crash.}
	\label{fig.4}
\end{figure*}

Figure \ref{fig.4} describes the network topology that captures the causal factors to produce a probability of a UAV crash due to failure. Observing from top to bottom, the observable nodes provide the information necessary to compute the prior probability of connected events. Each node has a set of discrete states (NO state; cause or fault is absent) and (YES state; cause or fault is present). For example, the state pilot error will have the value YES in case of pilot error and NO otherwise.
Data necessary to drive UAV risk model can be obtained based on expert knowledge, experiments or by learning from historical data. Thus, the next step involves the data collection.

\subsubsection{Data collection}

The Safety Management Document provided by ICAO \footnote{(2013) Safety management manual (doc 9859). [Online]. Available: www.icao.int/safety/SafetyManagement/Documents/Doc.9859.3rd\%20Edition.

alltext.en.pdf.} indicates that past accident and incident data collection is a key step and vital source of data, as they give plenty of information that can be helpful to validate the UAV risk analysis study.

{Due to limitation (or probably absence) of concrete data of drone crashes it is very challenging to derive exact probabilities of events \cite{NAP25143}. Thus, based on review of many references, we set some illustrative probabilities (or frequencies) to show how to apply the methodology, which is abstract and remain valid from theoretical perspective. { In case data is available in the future, the analyst will be able to derive exact and more realistic probabilities.}}

Consequently, in this paper, the data necessary to drive UAV risk model was collected through a review of previous research
papers \cite{oncu2014analysis, clothier2015safety, asim2005probable, belzer2017unmanned, susini2014technocritical}, multiple online accident and incident databases \footnote{(1960-2015) Causes of fatal accidents by decade. [Online]. Available:
http://www.planecrashinfo.com/cause.htm.}, and through a general website search \footnote{(2010-2016) UAV risk analysis. [Online]. Available: http://www.crash-aerien.news/forum/drones-uav-etudes-des-risques-t35137.html.}. This was done by posing this question: what are the common factors that lead to UAVs accidents and incidents ?.

Clothier and Walker \cite{clothier2015safety} used sample data from the U.S. Department of Defense. This study of military UAVs accidents and common failure categories identified common { pilot error as a cause of 17\% accidents, autopilot control module failure 26\% and loss of UAV electrical power 37\%.} The principal sources of accidents according to \cite{susini2014technocritical} was collected from an article published by the Washington Post, which reports that the annual number of crashes has risen over the past decade, to 26 crashes in 2012 and 21 in 2013, and the common cause of a crash was {loss of UAV electrical power for 38\%, autopilot control module failure 19\% and pilot error factor 17\%. }
 
The analysis in \cite{oncu2014analysis} was based on summaries of UAV incidents between 2010 and August 2014. This information was obtained from the Naval Safety Center in Norfolk, Virginia. Causal factors were related to{ pilot error factors (65\%), 16\% related to system/component failure or malfunction factors, and 19\% related to special factors}. In \cite{asim2005probable}, UAV accident data for the period 1995 - 2005 was taken from the U.S. Army accident database. These accidents are summarized under { system/component failure or malfunction (32\%), pilot error (11\%), and weather effect factors (5\%). }

 Belzer et al. \cite{belzer2017unmanned} analyse data from the UAS database of accidents and incidents between September 2001 and July 2016. The three most common errors were {flight crew/pilot error factors at 48\%, system/component failure or malfunction 31\% and mid-air collisions 19\%}. In \cite{wild2017post}, data were collected from a 10-year period, 2006 to 2015, sourced from the FAA Aviation Safety Information Analysis and Sharing System, NASA Aviation Safety Reporting System and the Civil Aviation Authority. The collected data indicate that {system/component failure or malfunction (63\%), pilot error (15\%) and navigation error (GPS loss) (11.53\%) were the most causal factor leading to UAV crashes.} Table V summarizes the percentage of common factors contributing towards the occurrence of UAV crash. 
 \begin{table*}[htb!]
 \caption{Percentage of causal factors contributing to UAV crash.}
 \begin{threeparttable}
\begin{tabular}{|c|c|c|c|c|c|c|c|c|c|c|c|}

\hline
{\color[HTML]{000000} REF} & {\color[HTML]{000000} \begin{tabular}[c]{@{}c@{}}Pilot\\  error\\ (PE)\end{tabular}} & {\color[HTML]{000000} \begin{tabular}[c]{@{}c@{}}System/\\ Component\\ Failure or\\  malfunction\\ (SCFM)\end{tabular}} & {\color[HTML]{000000} \begin{tabular}[c]{@{}c@{}}Weather\\  effects\\ on UAV\\ (WE)\end{tabular}} & {\color[HTML]{000000} \begin{tabular}[c]{@{}c@{}}GPS\\  loss (GL)\end{tabular}} & {\color[HTML]{000000} \begin{tabular}[c]{@{}c@{}}Air traffic\\ management\\  Failure\\ (ATMF)\end{tabular}} & {\color[HTML]{000000} \begin{tabular}[c]{@{}c@{}}Inability to\\ avoid\\  collision\\ (IAC)\end{tabular}} & {\color[HTML]{000000} \begin{tabular}[c]{@{}c@{}}Security\\  attacks\\ on the\\  drone \\ (SAOD)\end{tabular}} & {\color[HTML]{000000} \begin{tabular}[c]{@{}c@{}}Mid-air\\  collisions \\ (MC)\end{tabular}} & {\color[HTML]{000000} \begin{tabular}[c]{@{}c@{}}Degraded \\ communication \\ quality\\ (DCQ)\end{tabular}} & {\color[HTML]{000000} \begin{tabular}[c]{@{}c@{}}Autopilot\\  controller\\ module\\  failure\\ (ACMF)\end{tabular}} & {\color[HTML]{000000} \begin{tabular}[c]{@{}c@{}}Loss of UAV \\ electrical\\  power\\ (LEP)\end{tabular}} \\ \hline
\cite{oncu2014analysis} & 65 & 16 & 18 &  &  &  &  &  &  &  &  \\ \hline
\cite{asim2005probable} & 11 & 32 & 5 &  &  &  &  &  &  &  &  \\ \hline
\cite{belzer2017unmanned} & 48 & 31.2 & 3.6 & 13.99 & 1.70 & 0.567 & 0.18 & 19.65 &  &  & 10.77 \\ \hline
\cite{susini2014technocritical} & 17 &  &  &  &  &  &  &  & 14 & 19 & 38 \\ \hline
\cite{clothier2015safety} & 17 &  &  &  &  &  &  &  & 11 & 26 & 37 \\ \hline
\tnote{13} & 58 & 17 & 6 &  &  &  & 9 &  &  &  &  \\ \hline
\tnote{14} &  & 6.16 &  & 17.02 & 7.97 &  & 1.47 & 38.37 &  &  & 2.97 \\ \hline
\cite{wild2017post} & 15 & 63 & 4.84 & 11.53 & 5.38 & 5.42 &  & 1.55 &  &  &  \\ \hline
\end{tabular}
\begin{tablenotes}
  \item[13] (1960-2015) Causes of fatal accidents by decade. [Online]. Available:
http://www.planecrashinfo.com/cause.htm.
  \item[14] (2010-2016) Uav risk analysis. [Online]. Available: http://www.
crash-aerien.news/forum/drones-uav-etudes-des-risques-t35137.html.
  \end{tablenotes}
    \end{threeparttable}

\end{table*}

\begin{figure*}
	\centering
	\includegraphics[width=7in,height=3in]{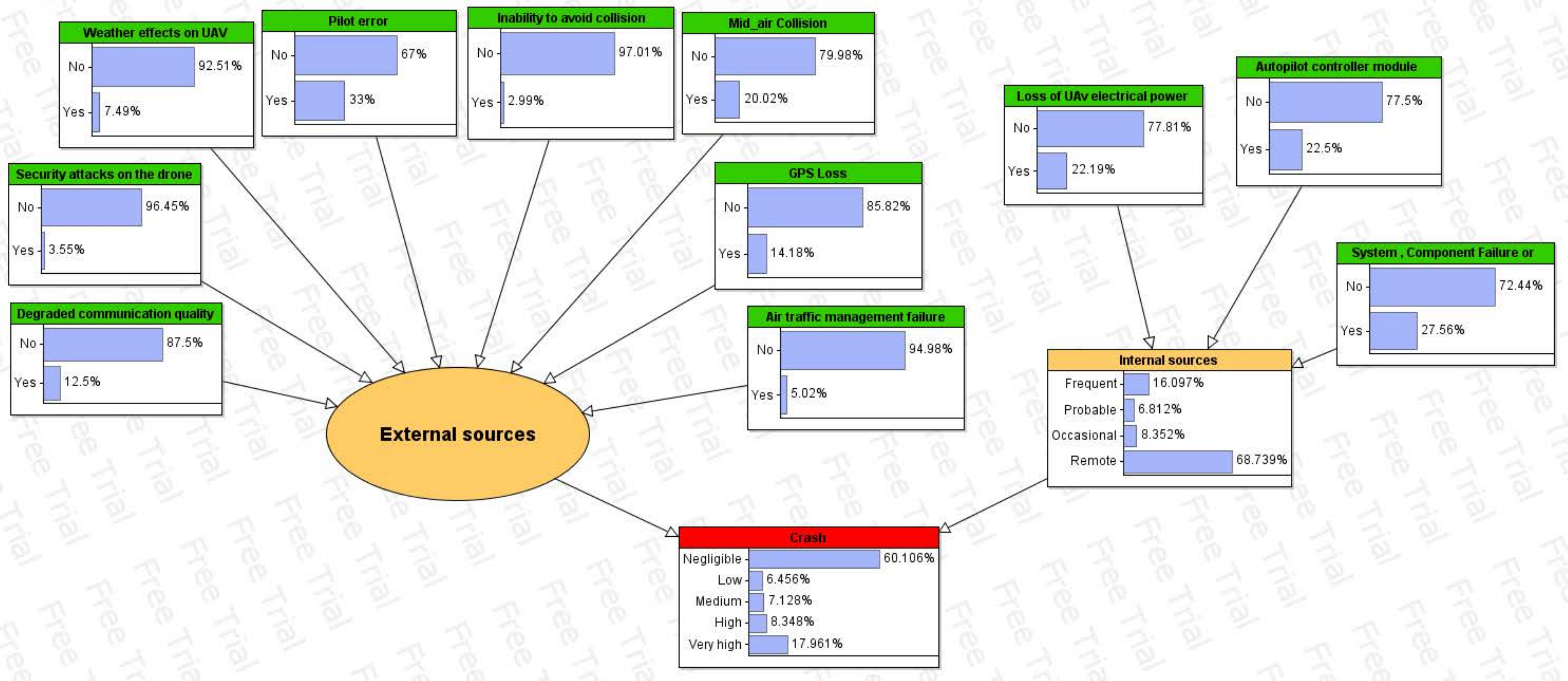}
	\caption{{{Joint frequency distribution of risk of crash.}}}
	\label{fig.5}
\end{figure*}

The prior distribution of the root causes of crash risk (observable nodes) were input to AgenaRisk \footnote{N. Birtles, N. Fenton, M. Neil, and E. Tranham, ``Agenarisk manual
(version 6.1) computer software,'' 2014.} (Bayesian Network and Simulation Software for Risk Analysis and Decision Support) to infer the posterior distribution of each node in BN. Based on the prior {contribution} of observable nodes, the joint {contribution} of the intermediate nodes could be obtained. The intermediate nodes have several parent nodes, so their CPT structure is large. AgenaRisk software was used to mitigate this difficulty, as it helps to calculate the conditional probabilities. {It is clarified that due to the lack of real probability values, which necessitate the knowledge of exposure rates per factor, we used the frequencies of each factor's presence in the UAV-related events, as explained above. Therefore, wherever the term probability is used, this actually denotes the frequency of contribution of each factor.} Using the parameters defined by AgenaRisk, coupled with weights among the nodes defined, probability values in CPT can be calculated rapidly. 
Figure \ref{fig.5} shows the Bayesian network topology and its initial states. Both intermediate nodes output a set of four ordered states: frequent, probable, occasional and remote. Using the AgenaRisk software, {a UAV crash due to the combination of the internal and external factors considered will occur with a frequency that is:} negligible (60.106\%), low 
(6.456\%), medium (7.128\%), high (8.348\%) and very
high (17.961\%) as shown in Fig. \ref{fig.5}.
\subsubsection{Using the BN network to reason with risk}

The inference algorithms types are predictions, diagnostics, combined and intercausal \cite{korbbayesian}. In our case, we were focused on how to use BN for prediction calculations in order to demonstrate how hazardous factors identified in previous sections can affect and change the frequency of occurrence of a crash given a fault {from the list of the causal factors considered in the study.}
\paragraph{Causal inference}
Causal inference estimates the posterior probability of a certain child node of the observed evidence. It is also called forward inference since the inference direction is from evidence to their child nodes. Causal inferences are made by a simple query. A simple query computes $P (X_{i}/e)$, where the evidence \textit{e} is the ancestor of $X_{i}$. A causal inference is suitable for finding the probability of a certain fault, or the most likely fault after updating the evidence \cite{jun2017bayesian}. As illustration, let us take the example of pilot error as evidence in the causal inference in order to predict what will happen. For example, after detecting the evidence of pilot error that {contributed to the crash}, which means a {frequency} of 100\% for the YES state, we can update and estimate the posterior {distribution of frequencies} of a crash based on that new evidence using the BN model, causally related to the pilot error. {This inference reveals the extent to which the causal evidence, i.e the pilot error evidence, affects the posterior {contribution to} the target node, i.e ``crash''.} {As a result, given that a pilot error is detected,} {the {frequency} distribution} {of a crash is changed: negligible (51.046\%), low (7.44\%), medium (8.283\%), high (10.161\%) and very high (23.071\%)}. The new {frequencies are} easily calculated using the AgenaRisk software by updating the state of the pilot error to be 0\% for FALSE state and 100\% for the TRUE state.
\paragraph{Diagnostic inference}
The diagnostic inference estimates the posterior probability of a certain parent node from the observed evidence. It is called backward inference, since the inference direction is from evidence to its parents. Like causal inferences, diagnostic inferences are made by a simple query. A simple query calculates $P (X_{i}/e)$, where the evidence \textit{e} is a child of $X_{i}$. Diagnostic inferences are appropriate for determining the magnitude of a cause-effect on symptoms \cite{jun2017bayesian}.
Different from causal inference, diagnostic inference estimates the posterior probability of a cause node, given an effect node as evidence. Let us take the crash as an evidence in the diagnostic inference. For example, we could think of a simple diagnostic inference query: $P (X_{CE} = 1|X_{FC} = 1)$. That is, we can estimate the posterior {frequency} of external sources under the evidence of crash. As the result of the query, the {distribution of frequencies} of external factors is frequent (42.76\%), probable (6.247\%), occasional (6.787\%) and remote (44.206\%), after finding the evidence of ``crash = very high''.

\subsubsection{Scenario analysis}
{This section provides two examples that demonstrate the use of the BN model presented in Fig. \ref{fig.4} to identify the parameters that have high impacts on the risk of crashes and estimate the {frequency distribution} of UAV crash.}

\paragraph{Scenario one: Hypotheses of external sources}
\begin{figure*}[htb!]
\centering\includegraphics[width=7in,height=3in]{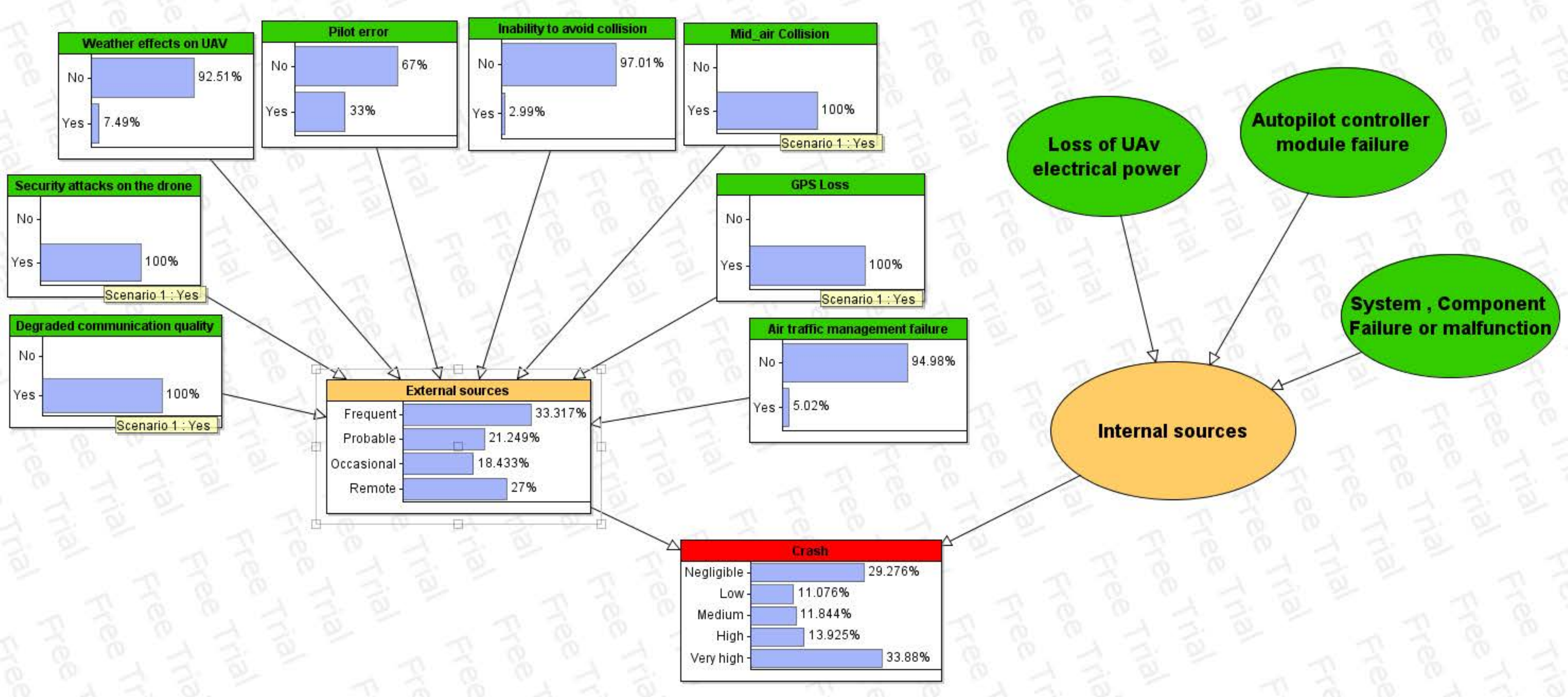}\caption{Scenario one where external sources are considered.}\label{fig.6}\end{figure*}
In this scenario, the UAV risk of crash under specific external conditions is estimated. Four external factors are chosen to represent the effect of the external source of hazards on the risk estimation. This scenario is demonstrated in Fig. \ref{fig.6}. {The distribution of UAV risk of crash is shown as negligible (29.276\%), low (10.076\%), medium (11.844\%), high (13.925\%) and very high (33.88\%), which clearly indicates a remarkable increase of risk level compared to the prior distribution. }This means, in case of the occurrence of an external error, the {frequency} level of a crash increases from 17.961\% to 33.88\%, which represents a very low
safety flight condition. Of course, this {frequency} is based on initial {frequency} distributions assumptions assigned to each observable state. In the case of real-time safety monitoring, {with the use of probability data and not frequencies as used here due to unavailability of the former, }when the probability of the crash is very high, the responsible agent will send a command to activate the risk mitigation techniques summarized in Table III and Table IV.

\paragraph{Scenario two: Hypotheses of internal sources}
\begin{figure*}
\centering\includegraphics[width=7in,height=3in]{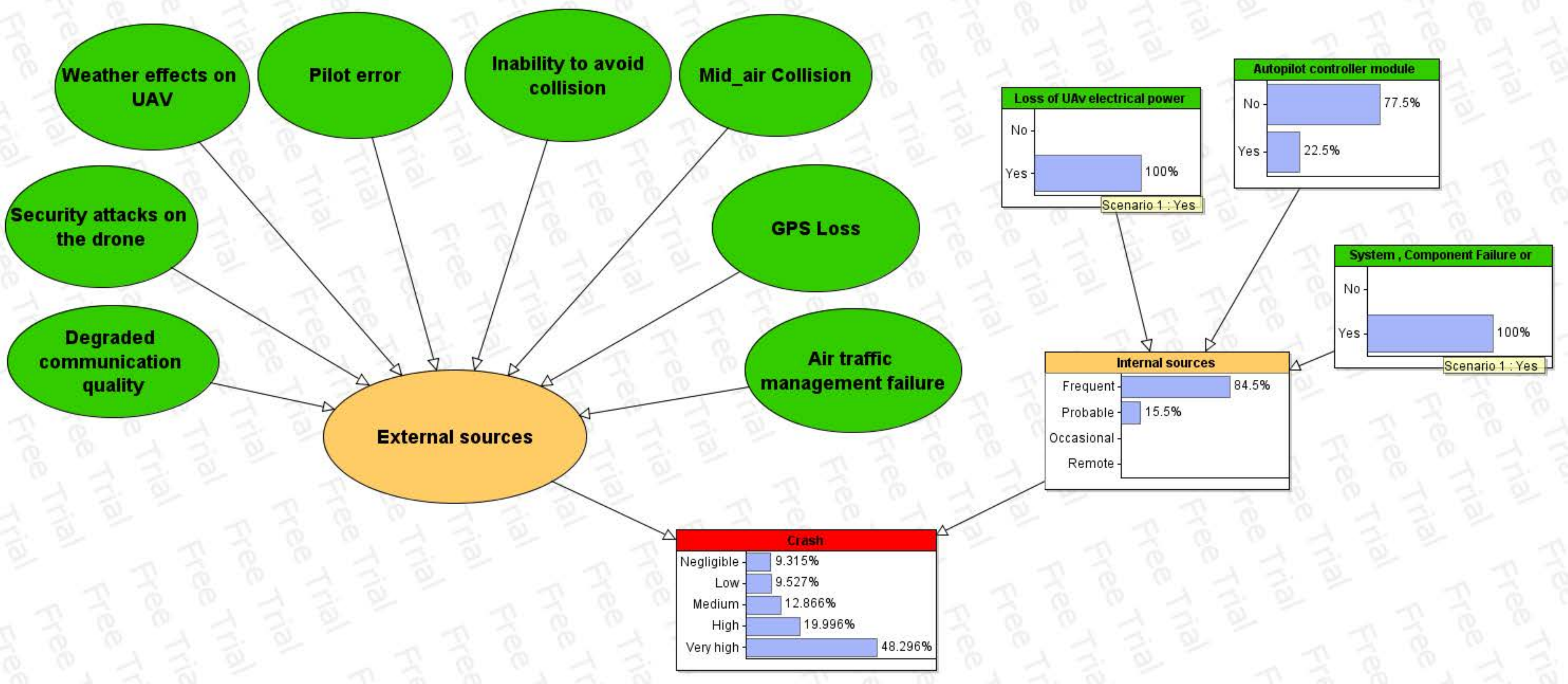}\caption{Scenario two where internal sources are considered.}\label{fig.7}\end{figure*}
In this scenario, two internal sources of hazards are considered. The results for this scenario are shown in Fig. \ref{fig.7}. {As per the result of the BN model, the distribution of UAV risk of crash in this scenario can be demonstrated as negligible (9.315\%), low (9.527\%), medium (12.866\%), high (19.996\%) and very high (48.296\%),} which is a significant increase in risk level. The large value of the ``very high'' state in this scenario indicates that certain risk control options must be used for these specific conditions in order to mitigate or reduce the {frequency} of crash. It also indicates that the risks of the crash would become high with only a small change in {the distribution of frequencies.}

\subsubsection{Sensitivity Analysis}
Sensitivity analysis plays an important role in probabilistic risk assessment, illustrating the performance of each risk factor\textquotesingle s contribution to the occurrence of crash accidents of UAV. Considering the BN model of Fig. \ref{fig.4} again, it is interesting to know what are the nodes that have the greatest {contribution to} the node ``crash''.

\begin{figure*}[htb!]
	\centering
		\includegraphics[width=7in,height=4in]{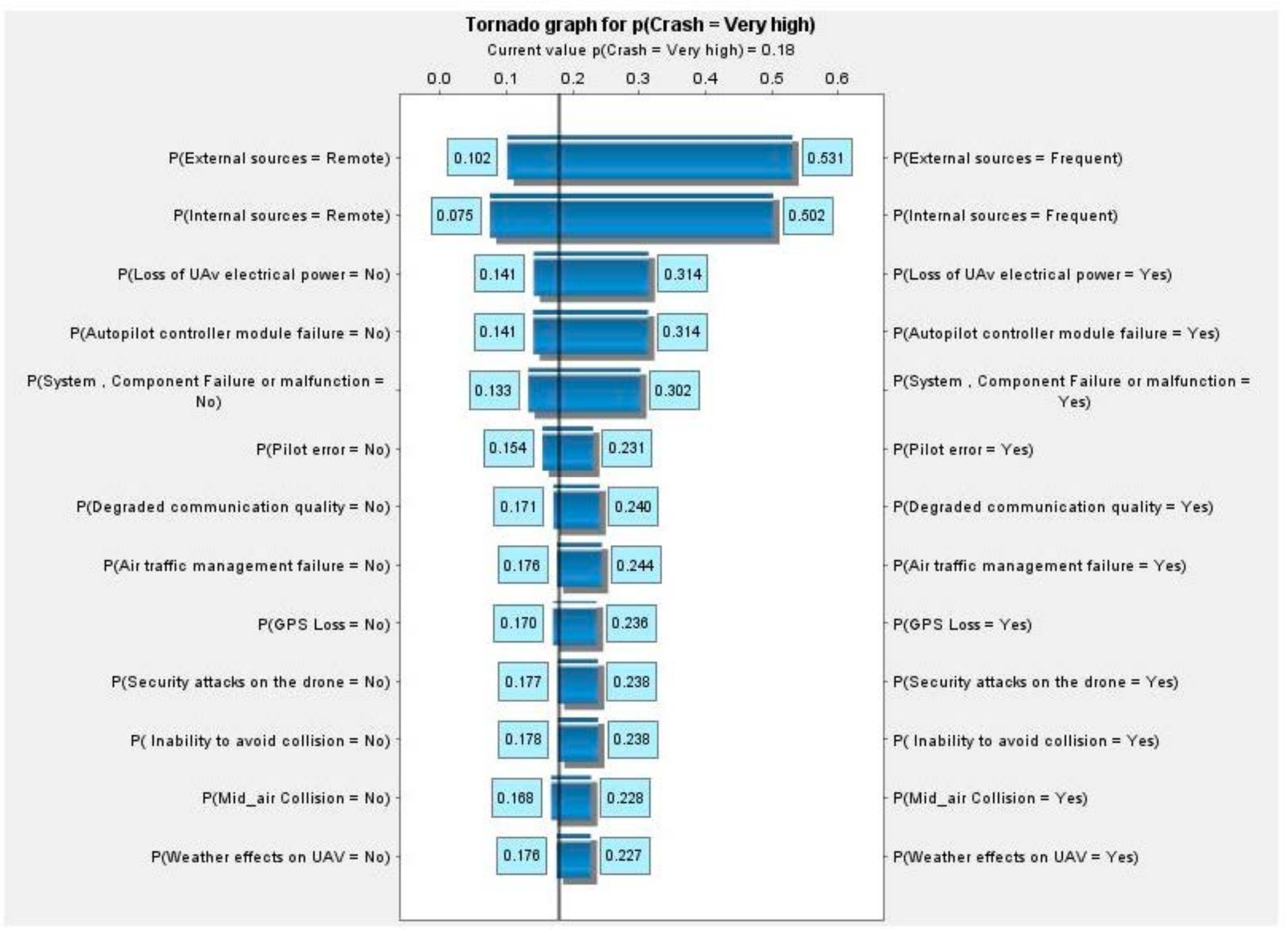}
	\caption{Tornado graph showing which nodes most {frequent} risk of crash.}
	\label{fig.8}
\end{figure*}

AgenaRisk does this automatically by allowing us to select a target node and any number of other nodes (called sensitivity nodes). So, setting Crash as the target node we automatically obtain the tornado graph in Fig. \ref{fig.8}. From a purely visual perspective, we can think of the length of the bars
corresponding to each sensitivity node in the tornado graph as being a measure of the {contribution} of that node {to }the target one.
Thus, the node external sources has by far the most {contribution to} the node ``crash''. The formal interpretation is that the {frequency} of crash given the result of external sources go from 0.102 (when external sources' occurrence is remote) to 0.531 (when external sources' occurrence is frequent). The {frequency} of crash given the result of internal sources go from 0.075 (when internal sources' occurrence is remote) to 0.502 (when internal sources' occurrence is frequent). The vertical bar on the graph is the marginal {frequency} for crash being very high (0.18).

\section{CONCLUSIONS}
This paper provides a functional safety methodology for drone crashes. Two approaches were used. The first approach followed a qualitative safety risk analysis based on international safety standards ISO 12100 and ISO 13849, in which an analysis was conducted to identify hazards, along with their possible causes existing safety mitigation methods and maintaining functional safety recommendations, associated with proposed UAV application and use case. The second approach used a probabilistic model-based risk analysis method. By using Bayesian Networks, a general crash accident model was derived, incorporating causal relationships and conditional probabilities. The use of Bayesian Networks {as} a modelling tool well suited to, test and visualize hazard scenarios, estimate the {frequency} of UAV crash and enable identification of the factors that have the greatest {contribution to} the occurrence of UAV crash. An example was given, and the simulation results showed the feasibility of the proposed methodology.

Future work will extend the experimental evaluations to include UAV simulations and real flight-testing as a means to use probability figures instead of frequencies that were used here for demonstration purposes. By the writer\textquotesingle s opinions and experiences, the processes of verification and
validation of safety functions are not easy to achieve
and get more effective result. Therefore, in order to get more
effective result, we plan to verify and validate the functional safety system to make sure that the safety system meets the functional requirements, and are suitable for the risk reduction.
We are also investigating how to implement artificial intelligence and cloud-based safety assurance modules to allow monitoring and ensuring the safe operation of drones during their mission.
\section*{Acknowledgments}
This work is supported by the Robotics and Internet-of Things (RIoT) Unit at Center of Excellence and the Research
Initiatives Center (RIC) of Prince Sultan University. It is also supported by Gaitech Robotics in China.

\bibliographystyle{IEEEtran}
\bibliography{IEEEabrv,biblio}

\begin{IEEEbiographynophoto}{\textbf{AZZA ALLOUCH}} was born in Sfax, Tunisia, in 1991. She received her master degree from the National School of Electronics and Telecommunication of Sfax, in 2016.
She is currently pursuing the Ph.D. degree with
the Computer Laboratory for Industrial Systems,
National Institute of Applied Science and Technology,
and the Faculty of Mathematical, Physical and Natural Sciences of Tunis.
Her interests focus on Machine learning, Blockchain and UAV.
\end{IEEEbiographynophoto}

\begin{IEEEbiographynophoto}{\textbf{ANIS KOUBAA}} received the M.Sc. degree from University Henri Poincar\'e, France, in 2001, and the Ph.D. degree from INPL, France, in 2004. He is currently a Professor of Computer Science, Aide to Rector of Research Governance, and the Director of the Robotics and Internet of Things Research Lab in Prince Sultan University. He is also a Senior Researcher with CISTER/INESC and ISEP-IPP, Porto,
Portugal, and a Research and Development Consultant
with Gaitech Robotics, China. His current research interests include providing solutions towards the integration of robots and drones into the Internet of Things (IoT) and clouds, in the context of cloud robotics, robot operating system, robotic software engineering, wireless communication for the IoT, real-time communication, safety and security for cloud robotics, intelligent algorithms design for mobile robots, and multi-robot task allocation. He is also a Senior Fellow of the Higher Education Academy, U.K. He has been the Chair of the ACM Chapter, Saudi Arabia, since 2014.

\end{IEEEbiographynophoto}

\begin{IEEEbiographynophoto}{\textbf{MOHAMED KHALGUI}} received the B.S. degree
in computer science from Tunis El Manar University,
Tunis, Tunisia, in 2001, the M.S. degree
in telecommunication and services from Henri
Poincar\'e University, Nancy, France, in 2003,
the Ph.D. degree from the National Polytechnic
Institute of Lorraine, Nancy, in 2007, and the
Habilitation Diploma degree in information technology
(computer science) from the Martin Luther
University of Halle-Wittenberg, Halle, Germany,
in 2012, with Humboldt Grant.
He was a Researcher in computer science with the Institut National
de Recherche en Informatique et Automatique, Rocquencourt, France,
the ITIA-CNR Institute, Vigevano, Italy, the Systems Control Laboratory,
Xidian University, Xi'an, China, and the KACST Institute, Riyadh,
Saudi Arabia, a Collaborator with SEG Research Group, Patras University,
Patras, Greece, the Director of the RECS Project, O3NEIDA, Canada,
the Director of the RES Project, Synesis Consortium, Lomazzo, Italy,
the Manager of the Cyna-RCS Project, Cynapsys Consortium, France,
and the Director of the BROS and RWiN Projects, ARDIA Corporation,
Germany.
He is currently a Professor with Jinan University, China. He has been
involved in various international projects and collaborations. He is a TPC
member of many conferences and different boards of journals.
\end{IEEEbiographynophoto}

\begin{IEEEbiographynophoto}{\textbf{TAREK ABBES}}
is an assistant professor at the National School of Electronics and Telecommunication of Sfax, Tunisia. He received his Ph.D diploma in computer science from Henri Poincar\'e University Nancy, France in 2004. The topic of his thesis was ``network traffic classification and multipattern matching for intrusion detection''. He obtained the DEA Diploma in Artificial Intelligence from Henri Poincar\'e University of Nancy France in 2001 and a Telecommunication engineering Diploma from High school of communication of Tunis in 2000.
His research is concerned with formal verification of network equipments configuration, network intrusion detection and feature interaction in Telecommunication systems.
\end{IEEEbiographynophoto}


\end{document}